\documentclass[journal]{IEEEtran}

\usepackage{amsmath,amssymb,amsfonts}
\usepackage{soul}
\usepackage{textcomp}
\usepackage[table]{xcolor}
\usepackage{algorithm}
\usepackage{algpseudocode}
\usepackage{subfigure}

\DeclareMathOperator*{\argmin}{\arg\min}  
\newcommand{\Mod}[1]{\ (\mathrm{mod}\ #1)}
\usepackage{framed}
\usepackage{float}
\usepackage{hyperref}

\usepackage[pdftex]{graphicx}

\ifCLASSOPTIONcompsoc
  \usepackage[nocompress]{cite}
\else
  \usepackage{cite}
\fi

\makeatletter
\def\ps@IEEEtitlepagestyle{
  \def\@oddfoot{\mycopyrightnotice}
  \def\@evenfoot{}
}
\def\mycopyrightnotice{
  {\footnotesize
  \begin{minipage}{\textwidth}
  \centering
  Copyright~\copyright~2015 IEEE. Personal use of this material is permitted. However, permission to use this  \\ 
  material for any other purposes must be obtained from the IEEE by sending a request to pubs-permissions@ieee.org.
  \end{minipage}
  }
}

\begin{document}

\title{An Efficient and Reliable Asynchronous Federated Learning Scheme for Smart Public Transportation}

\author{Chenhao~Xu,
        Youyang~Qu,~\IEEEmembership{Member,~IEEE,}
        Tom H. Luan,~\IEEEmembership{Senior~Member,~IEEE,}
        Peter W. Eklund,
        Yong~Xiang,~\IEEEmembership{Senior~Member,~IEEE,}
        and~Longxiang~Gao,~\IEEEmembership{Senior~Member,~IEEE}%
\thanks{Longxiang Gao is the corresponding author.}%
\thanks{Chenhao Xu and Yong Xiang are with the Deakin Blockchain Innovation Lab, School of Information Technology, Deakin University, Geelong, Australia. E-mail: \{xueri and yong.xiang\}@deakin.edu.au.}%
\thanks{Youyang Qu is with Data 61 Australia Commonwealth Scientific and Industrial Research Organization, Australia. E-mail: youyang.qu@data61.csiro.au.}%
\thanks{Tom H. Luan is with School of Cyber Engineering, Xidian University, Shaanxi, China. E-mail: tom.luan@xidian.edu.cn.}%
\thanks{Peter W. Eklund is with School of Information Technology, Deakin University, Geelong, Australia. E-mail: peter.eklund@deakin.edu.au.}%
\thanks{Longxiang Gao is with Qilu University of Technology and Shandong Computer Science Center, China. E-mail: gaolx@sdas.org.}%
}

\markboth{IEEE Transactions on Vehicular Technology}%
{Chenhao \MakeLowercase{\textit{et al.}}: An Efficient and Reliable Asynchronous Federated Learning Scheme for Smart Public Transportation}

\maketitle

\begin{abstract}
Since the traffic conditions change over time, machine learning models that predict traffic flows must be updated continuously and efficiently in smart public transportation.
Federated learning (FL) is a distributed machine learning scheme that allows buses to receive model updates without waiting for model training on the cloud.
However, FL is vulnerable to poisoning or DDoS attacks since buses travel in public.
Some work introduces blockchain to improve reliability, but the additional latency from the consensus process reduces the efficiency of FL. 
Asynchronous Federated Learning (AFL) is a scheme that reduces the latency of aggregation to improve efficiency, but the learning performance is unstable due to unreasonably weighted local models.
To address the above challenges, this paper offers a blockchain-based asynchronous federated learning scheme with a dynamic scaling factor (DBAFL).
Specifically, the novel committee-based consensus algorithm for blockchain improves reliability at the lowest possible cost of time.
Meanwhile, the devised dynamic scaling factor allows AFL to assign reasonable weights to stale local models.
Extensive experiments conducted on heterogeneous devices validate outperformed learning performance, efficiency, and reliability of DBAFL.
\end{abstract}


\IEEEpeerreviewmaketitle

\section{Introduction}
\label{sec:introduction}

\IEEEPARstart{M}{achine} learning (ML) is a popular approach on the Internet of Vehicles (IoV) to enable smart public transportation~\cite{lu2020blockchain, kong2021fedvcp}.
For example, buses with ML models are able to forecast traffic flows and the time passengers wait at stops, assisting drivers in improving driving safety and fuel economy. However, because traffic conditions change over time, ML models must be updated continuously and efficiently.
Federated Learning (FL) is a distributed ML scheme that allows models to be trained locally and updated frequently.
For instance, buses collect traffic data and train local models, while roadside units (RSUs) periodically aggregate the local models to produce an accurate global model and send it back to the buses~\cite{mcmahan2017communication}.

However, FL raises efficiency and reliability concerns due to the limited computing resources and continuous movement of buses in smart public transportation. Firstly, the synchronous aggregation strategy forces the aggregation server to collect enough local models before aggregation, which is inefficient due to the difference of computing power and dataset sizes among buses and RSUs~\cite{xu2021asynchronous}. Specifically, high-performance nodes have to wait for lagging nodes to finish their training before aggregation. Secondly, unreliable local models gathered from buses pose FL at risk of poisoning attacks~\cite{lyu2020towards}. The centralized aggregation server of FL is subject to DDoS attacks~\cite{lu2020blockchain}. Both of these attacks reduce the reliability of FL.

To improve the efficiency of FL, asynchronous federated learning (AFL) is proposed, which reduces the latency by performing aggregation whenever a local model is received~\cite{chen2020asynchronous, chen2019communication, wu2020safa, lu2020privacy, deng2020adaptive}. However, due to the high aggregation frequency, there are outdated global models in AFL, from which stale local models trained usually have relatively low accuracy~\cite{xu2021bafl}. Existing work either discards or assigns irrational weights to stale local models, leading to unstable FL learning performance (i.e. convergence speed and global model accuracy)~\cite{xu2021asynchronous}.

Some work adopts blockchain to improve the reliability of FL~\cite{ali2021integration}. Due to its decentralized storage and attack-proof consensus algorithm~\cite{qu2020blockchained, xu2021light}, blockchain allows FL to conduct a decentralized and transparent training process, resulting in improved security and trustability. However, consensus algorithms of the blockchain are either compute-intensive (i.g. PoW) or communication-intensive (i.g. PBFT)~\cite{xu2021asynchronous}. To improve efficiency, committee-based consensus algorithms such as DPoS are proposed~\cite{bamakan2020survey}, but their token or reputation systems are unsuitable for buses that pass quickly. Although several blockchain-based AFL schemes are proposed~\cite{liu2021blockchain1, feng2021blockchain, yuan2021chainsfl}, their consensus processes are still time-consuming.

In order to address the above challenges and better apply FL into smart public transportation systems, this paper offers a blockchain-based asynchronous federated learning framework with a dynamic scaling factor (DBAFL). A novel committee-based consensus algorithm is introduced to improve reliability while bringing the least amount of communication burden. Specifically, the committee leader, as the aggregation server, identifies low-accuracy local models based on its local dataset to resist poisoning attacks. Without the need for communication and voting, a new committee leader is elected from RSUs periodically based on the hash of the latest block to reduce the risk of being subjected to DDoS attacks. Besides, when performing aggregation, a dynamic scaling factor is designed to assign appropriate weights to local models according to their accuracy and correspondingly improves the learning performance of FL. Experiments conducted on heterogeneous devices evaluate the proposed scheme and demonstrate its outstanding learning performance, efficiency, and reliability. The main contributions of this paper are as follows.

\begin{itemize}
    \item A blockchain-based asynchronous federated learning scheme is designed for smart public transportation, considering learning performance, efficiency, and reliability in heterogeneous computing environments.
    \item A dynamic scaling factor is designed to assign appropriate weights to stale local models with the joint effort of a committee-based consensus algorithm, allowing FL to efficiently converge to higher accuracy while being highly attack-resistant.
    \item An open-source prototype\footnote{The Github link is \url{https://github.com/xuchenhao001/AFL}.} is implemented with comprehensive experiments conducted to validate the advantages from three perspectives compared with state-of-the-art schemes.
\end{itemize}

The remainder of this paper is organized as follows: Section~\ref{sec:related_work} presents related work. Section~\ref{sec:model} models the proposed scheme in detail. Section~\ref{sec:analysis} analyzes the proposed scheme from several aspects. Section~\ref{sec:system_evaluation} evaluates the proposed scheme experimentally. Finally, Section~\ref{sec:summary_and_future_works} summarizes the paper and outlines future work.

\section{Related Work}
\label{sec:related_work}

The related work of this paper includes the blockchain, federated learning, IoV, and asynchronous federated learning.

\subsection{Blockchain, Federated Learning, and IoV}

Some existing work integrates the blockchain and FL to resist attack in classic FL~\cite{qu2020blockchained, ur2020towards, weng2019deepchain, peng2021vfchain, shayan2020biscotti, li2020blockchain, chai2020hierarchical, pokhrel2020federated, lu2020blockchain, kang2020reliable, liu2021blockchain1, feng2021blockchain, yuan2021chainsfl}. The blockchain is a motivation for the nodes to participate in FL and contribute high-quality local models~\cite{ur2020towards, weng2019deepchain, kang2021optimizing}, a distributed reputation management system to resist repudiation and tampering~\cite{kang2020reliable}, and an auditable distributed database that allows FL to conduct a transparent training process~\cite{peng2021vfchain}. However, none of these schemes simultaneously take efficiency and reliability into account. Shayan \emph{et al.}~\cite{shayan2020biscotti} prove that the blockchain effectively defends FL against poisoning attacks. Li \emph{et al.}~\cite{li2020blockchain} design a committee consensus algorithm for blockchain-based FL without analyzing the communication burden brought by the blockchain. Besides, the score-based committee election in their scheme is unsuitable for fast-traveling buses. Kang \emph{et al.}~\cite{kang2020scalable} propose a hierarchical blockchain-based FL scheme with improved efficiency and privacy, but miner election and model quality cross-validation in the consensus process are time-consuming and inappropriate for buses.

There are some work adopts FL in IoV. For example, Lim \emph{et al.}~\cite{lim2021towards} propose a blockchain-based IoV network that matches the lowest cost Unmanned Aerial Vehicles (UAVs) to each subregion, but communication latency is not analyzed and tested.
Besides, some work utilizes the blockchain to ensure a secure FL framework on IoV networks, including data sharing~\cite{chai2020hierarchical}, driving assistance~\cite{chen2021bdfl}, and intrusion detection~\cite{liu2021blockchain}. However, these schemes are inefficient on IoV networks due to the synchronous aggregation strategy. Pokhrel \emph{et al.}~\cite{pokhrel2020federated} optimize the latency by adjusting the block arrival rate, but the PoW consensus in their scheme is still time-consuming. Lu \emph{et al.}~\cite{lu2020blockchain} introduce the directed acyclic graph architecture to the blockchain to improve efficiency, but security and communication latency are not analyzed.

DBAFL introduces a novel committee-based consensus algorithm to the blockchain, which brings the least communication latency to buses while ensuring reliability.

\subsection{Asynchronous Federated Learning}

FL, proposed in 2017~\cite{mcmahan2017communication}, is a distributed learning scheme applied in various scenarios~\cite{nishio2019client, khan2020federated, qu2020decentralized, xu2021scei}. To reduce aggregation latency and improve efficiency on resource-limited networks, AFL is proposed~\cite{chen2020asynchronous, chen2019communication, liu2021blockchain1, wu2020safa, lu2020privacy, deng2020adaptive}.

A semi-asynchronous FL scheme is proposed for mitigating model staleness~\cite{wu2020safa}. However, lagging models in their scheme are given the same weight as normal models during aggregation.
Chen \emph{et al.}~\cite{chen2020asynchronous} assign a higher weight to stale local models due to large training datasets. But stale local models in IoV networks may be due to inadequate computing power.
Chen \emph{et al.}~\cite{chen2019communication} improve efficiency by reducing the updating frequency of parameters in deep layers, which is hard to apply to other types of models.
Liu \emph{et al.}~\cite{liu2021blockchain1} propose an AFL framework with a staleness coefficient to adjust the weight of stale local models, but analysis and validation are missing.
Lu \emph{et al.}~\cite{lu2020privacy} present a new gradient compression approach to improve efficiency at the cost of global model accuracy.
Deng \emph{et al.}~\cite{deng2020adaptive} propose a semi-AFL approach APFL that mixes the parameters of local and global models to lower communication frequency, but an additional training phase is required. 
Chen \emph{et al.}~\cite{chen2021bdfl} present a blockchain-based AFL scheme BDFL with all models saved in the blockchain and aggregated with the same weight. 
As highly relevant work, APFL and BDFL are included as benchmarks in Section~\ref{sec:system_evaluation}.

A dynamic scaling factor is designed in DBAFL for weighted aggregation according to model accuracy, which improves learning performance and reliability.

\section{DBAFL Modeling}
\label{sec:model}

This section explains the model of DBAFL in detail from the aspect of the system model, workflow, dynamic scaling factor, and committee-based consensus algorithm.

\subsection{System Model}

\begin{figure}[htp]
    \centering
    \includegraphics[width=\linewidth]{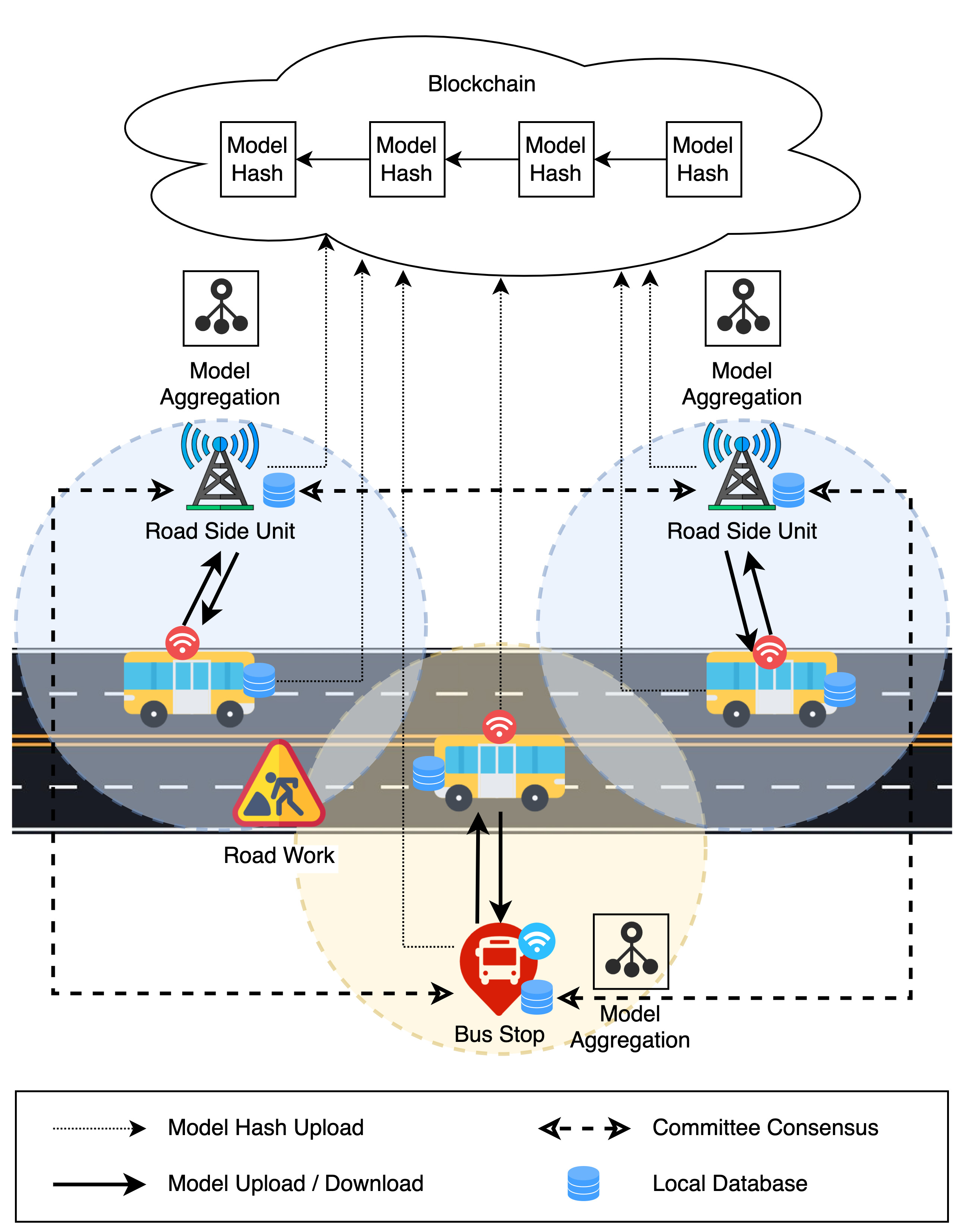}
    \caption{The architecture of DBAFL in public transportation.}
    \label{fig:architecture}
\end{figure}

The public transportation system includes buses and RSUs on a road, as shown in Fig.~\ref{fig:architecture}. The bus stop equipped with sensors and an edge computing server is considered a kind of RSU in this scenario. RSUs usually have higher computing and communication power than buses. Buses establish short-term Vehicle-to-RSU (V2R) and Vehicle-to-Vehicle (V2V) connections with nearby RSUs or buses when they are within a signal region, allowing them to transmit a certain amount of data. Besides, RSUs have a high-speed Ethernet connection with each other to support smart public transportation.

Buses and RSUs train ML models collaboratively to predict traffic flows and the time passengers wait at stops, allowing buses to travel safely and efficiently. Since traffic conditions change over time, buses and RSUs continuously collect training dataset through their sensors to update models. According to~\cite{lu2020blockchain, liu2021blockchain}, in an area, the traffic data collected on buses and RSUs are assumed independent and identically distributed (IID). However, the training time and model quality of buses and RSUs differ due to differences in computing power and dataset sizes, posing learning performance challenges to FL. The limited connect time and bandwidth of V2R and V2V connections pose efficiency challenges to FL. The poisoning and DDoS attacks launched by attackers on the roadside pose reliability challenges to FL.

In DBAFL, a dynamic scaling factor and a lightweight consensus algorithm are designed on top of AFL and the blockchain to improve learning performance, efficiency, and reliability. Specifically, buses and RSUs act as local nodes of FL and train local models based on their local data. After training, local nodes upload the hash value of the local model to the blockchain for other nodes to verify. Apart from that, buses upload the local model to nearby RSUs for global model aggregation. During the aggregation, the dynamic scaling factor assigns weights to local models according to their accuracy. RSUs are committee members and are eligible to be elected as the committee leader by the consensus algorithm. The identity of the next committee leader is determined by the hash of the most recent block, which does not require voting or communication. The committee maintains a distributed database with data synchronization to share models. The committee leader is in charge of performing aggregation whenever a new local model is received.

\subsection{Workflow}

\begin{figure}[htp]
    \centering
    \includegraphics[width=\linewidth]{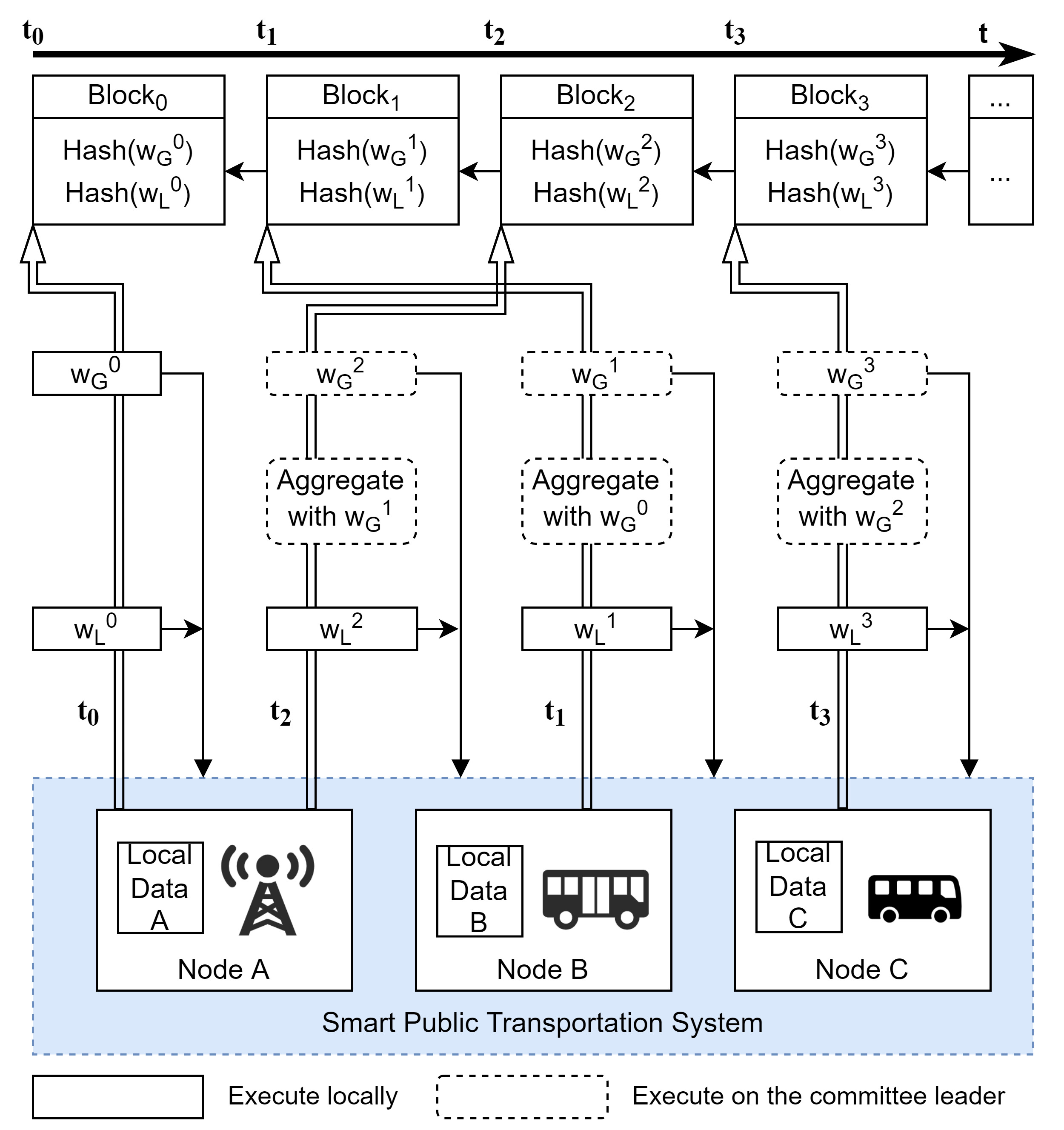}
    \caption{The workflow of DBAFL.}
    \label{fig:workflow}
\end{figure}

The workflow of DBAFL is illustrated in Fig.~\ref{fig:workflow}. In a smart public transportation system, assume three nodes (Node A, B, and C) are involved in the training process. Specifically, Node A is an RSU, while Node B and Node C are buses. With the passage of time (from $t_0$ to $t_3$), the nodes keep training local models with the latest global model. The details are as follows:

At $t_0$, Node A trains a local model $w_L^0$ as the initial global model $w_G^0$. Then Node A uploads the hash values of them, i.e. $\text{Hash}(w_G^0)$ and $\text{Hash}(w_L^0)$, to the blockchain to build up a genesis block $\text{Block}_0$ (The first block in the blockchain). After receiving the hash value of $\text{Block}_0$, each node downloads $w_G^0$ from Node A, and trains local models based on its local data. At the same time, a committee leader is elected based on the hash value of $\text{Block}_0$.

At $t_1$, Node B finishes its training and generates a new local model $w_L^1$. Next, Node B calculates the hash value $\text{Hash}(w_L^1)$ and uploads it to the blockchain. Besides, Node B uploads $w_L^1$ to the nearby RSU, where it will be stored in the distributed database and shared with the committee leader. Without waiting for other nodes, the committee leader aggregates $w_G^0$ and $w_L^1$ according to a dynamic scaling factor, which is explained in Section~\ref{sec:dynamic_scaling_factor}, to produce a new global model $w_G^1$. The committee leader then uploads $\text{Hash}(w_G^1)$ to the blockchain, where a new block $\text{Block}_1$ is generated. The global model $w_G^1$ is then shared with all committee members.

At $t_2$, Node A finishes training based on $w_G^0$ and acquires a stale local model $w_L^2$. Node A uploads $\text{Hash}(w_L^2)$ to the blockchain, along with $w_L^2$ shared with the committee leader through the distributed database. Based on $w_G^1$ and $w_L^2$, the committee leader generates a new global model $w_G^2$, along with the hash value $\text{Hash}(w_G^2)$ uploaded to the blockchain. After that, $\text{Block}_2$ is created and appended to the blockchain.

After a long period of training, Node C finally finishes training its local model $w_L^3$ based on $w_G^0$. Although it is stale, the local model $w_L^3$ uploaded by Node C is also acceptable for the committee leader. Node C uploads $\text{Hash}(w_L^3)$ to the blockchain at the same time. After aggregating $w_L^3$ and $w_G^2$, the committee leader generates a new global model $w_G^3$, and uploads $\text{Hash}(w_G^3)$ to the blockchain. Then, $\text{Block}_3$ is generated.

With the help of blockchain, the training process is consistent, transparent, and trustable. Besides, the novel committee-based consensus algorithm in the blockchain enables an attack-resistant DBAFL with a generalized ML model. The advantages of DBAFL are explained carefully in Section~\ref{sec:reliability}.

\subsection{Dynamic Scaling Factor}
\label{sec:dynamic_scaling_factor}

In~\cite{chen2019communication, chen2020asynchronous, liu2021blockchain1}, the authors demonstrate that the weight of local models during aggregation has a significant impact on federated learning performance. If a stale local model has low accuracy due to the limited computing resources of the node, relying too much on this particular local model results in a deterioration of global model accuracy. On the contrary, if the stale local model has relatively high accuracy due to large volumes of data on the node, it is preferential to make better use of it to improve the convergence speed of the global model.

A dynamic scaling factor, denoted as $\epsilon$, is designed to assign the newly arrived local model appropriate weight. To achieve a global model with high accuracy, the committee leader tests the accuracy of the models based on its local data before aggregation in DBAFL. Assuming $\mathcal{A}_L^t$ and $\mathcal{A}_G^{t-1}$ denote the test accuracy of the local model at time $t$ and that of the global model at time $t-1$, respectively, $\epsilon^t$ is defined as
\begin{equation}
\begin{gathered}
\epsilon^{t} = \frac{\mathcal{A}_L^t}{\mathcal{A}_G^{t-1}}.
\label{eq:scaling_factor}
\end{gathered}
\end{equation}
Considering the dynamic scaling factor $\epsilon^t$, the new global model at time $t$ is calculated by
\begin{equation}
\begin{gathered}
w_G^t = \frac{w_G^{t-1} + \epsilon^t \times w_L^t}{1 + \epsilon^t}.
\label{eq:aggregate}
\end{gathered}
\end{equation}
From Eq.~\ref{eq:aggregate}, it is straightforward that a greater value of $\epsilon$ means a higher weight of the local model. If the local model has higher accuracy than the global model, the committee leader increases $\epsilon$ to assign the local model with higher weight, and vice versa.

Since the committee leader is re-elected on a regular basis, as explained in Section~\ref{sec:consensus_algorithm}, assessing the model accuracy using local data by the committee leader is neutral and efficient while preserving data privacy. For example, as shown in Fig.~\ref{fig:consensus_algorithm}, after receiving the global model $w_G^0$ from RSU 1 and the local model $w_L^1$ from RSU 2, the committee leader, RSU 3, tests the accuracy of $w_G^0$ and $w_L^1$ based on its local data. Due to the disparity in data samples across nodes, the model accuracy assessing on RSU 3 is fairer than on RSU 1 or RSU 2. Besides, to strengthen the generality of the global model, a higher committee leader election frequency could be used.

\begin{algorithm}
\caption{Dynamic Scaling Factor}
\label{alg:async-fl}
\begin{algorithmic}[1]

\Function{Initialization}{} \Comment{\emph{On the first RSU}}
  \State initialize $w_L^0$ as $w_G^0$
  \State upload $\text{Hash}(w_L^0)$ and $\text{Hash}(w_G^0)$ to the blockchain
  \State save $w_L^0$ and $w_G^0$ to the distributed database
\EndFunction
\\
\Function{ClientUpdate}{} \Comment{\emph{On local nodes}}
  \For {each local epoch in \emph{E}}
    \State download $w_G^{t-1}$ from the nearby RSU
    \State $w_L^t$ $\leftarrow$ LocalTrain($w_G^{t-1}$ , localTrainData)
    \State upload $\text{Hash}(w_L^t)$ to the blockchain
    \State upload $w_L^t$ to the nearby RSU
  \EndFor
\EndFunction
\\
\Function{Aggregation}{} \Comment{\emph{On the committee leader}}
  \State wait $w_L^t$ in the distributed database
  \State $\mathcal{A}_L^t$ $\leftarrow$ LocalTest($w_L^t$, localTestData)
  \State $\mathcal{A}_G^{t-1}$ $\leftarrow$ LocalTest($w_G^{t-1}$, localTestData)
  \State $\epsilon^t$ $\leftarrow$ $\mathcal{A}_L^t/\mathcal{A}_G^{t-1}$
  \State $w_G^t = (w_G^{t-1} + \epsilon^t \times w_L^t) / (1 + \epsilon^t)$
  \State upload $\text{Hash}(w_G^t)$ to the blockchain
  \State save $w_G^t$ to the distributed database
\EndFunction
\end{algorithmic}
\end{algorithm}

Algorithm~\ref{alg:async-fl} shows the implementation details of the dynamic scaling factor in DBAFL. There are three functions in DBAFL: initialization (Lines 1 to 5), client update (Lines 7 to 14), and aggregation (Lines 16 to 24). The initialization process runs on the first RSU of the smart public transportation system, which trains a local model $w_L^0$ as the first global model $w_G^0$, as shown in Line 2. After uploading hash values of the models to the blockchain, the original models are stored in the distributed database on RSUs, as shown in Lines 3 and 4.

Before training its new local model $w_L^t$, each local node downloads the latest global model $w_G^{t-1}$ from the nearby RSU, as shown in Lines 9 and 10. After training, each local node uploads the hash value of the local model $\text{Hash}(w_L^t)$ to the blockchain and uploads the original local model $w_L^t$ to the nearby RSU, as shown in Lines 11 and 12. The loop continues if the local epoch $E$ is not reached, as shown in Lines 8. Note that the time $t$ in the client update function (running on local nodes) does not one-to-one correspond to the one in the aggregation function (running on the committee leader), due to the asynchronous aggregation strategy. $w_G^{t-1}$ refers to the latest global model downloaded from the nearby RSU.

From the view of the committee leader, the aggregation progress starts when a new local model $w_L^t$ is uploaded to the distributed database, as shown in Line 17. After testing based on the local data, the accuracies of $w_L^t$ and $w_G^{t-1}$ are obtained, as shown in Lines 18 and 19. Then, the dynamic scaling factor $\epsilon^t$ is calculated according to Eq.~\ref{eq:scaling_factor}, as shown in Line 20. Following that, a new global model $w_G^t$ is calculated according to Eq.~\ref{eq:aggregate}, as shown in Line 21. Finally, the committee leader uploads the hash value of the global model $\text{Hash}(w_G^t)$ to the blockchain and save the original global model $w_G^t$ to the distributed database, as shown in Lines 22 and 23.

\subsection{Committee-Based Consensus Algorithm}
\label{sec:consensus_algorithm}

\begin{figure*}[htbp]
    \centering
    \includegraphics[width=0.9\linewidth]{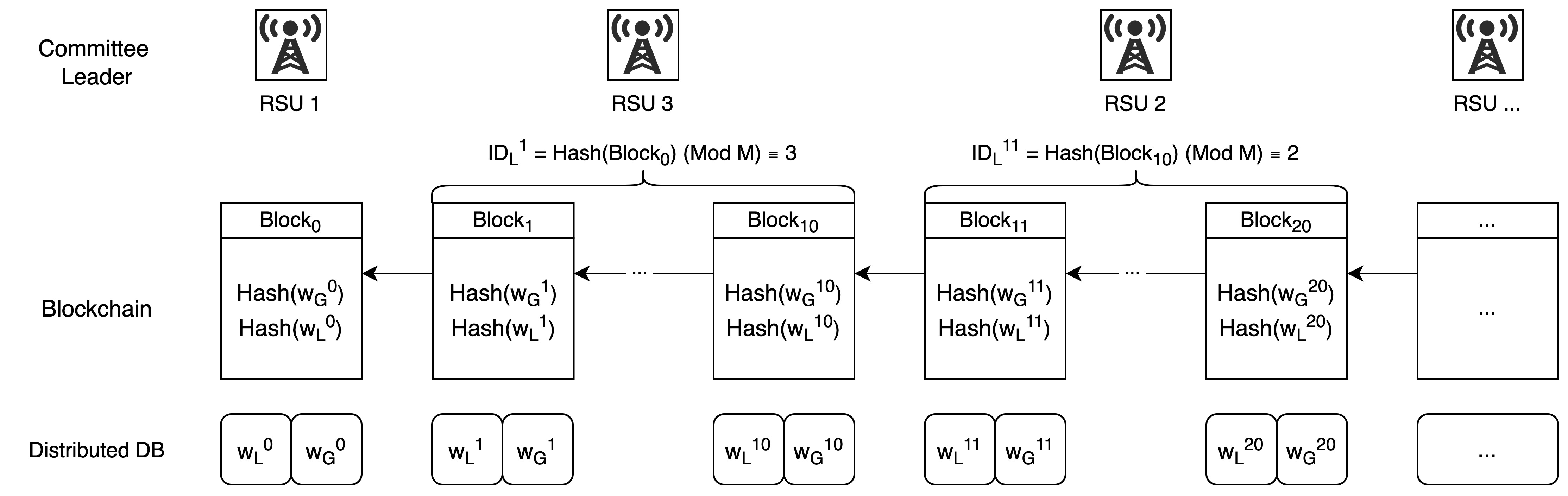}
    \caption{The committee leader is decided by the hash value of the latest block. The hash values of the models are stored in the blockchain while the original models are stored in distributed databases on RSUs. The relationship between the models in distributed databases and hash values on the blockchain is demonstrated.}
    \label{fig:consensus_algorithm}
\end{figure*}

In DBAFL, a new committee leader is elected after several new blocks are generated. Specifically, the frequency of electing new committee leaders increases as the reliability requirements grow. The identity of the new committee leader is calculated by the hash value of the latest block.

As shown in Fig.~\ref{fig:consensus_algorithm}, assuming the genesis block $\text{Block}_0$ is generated on RSU 1, the consensus algorithm in the blockchain ensures the replicated blocks on the other nodes are identical as the original $\text{Block}_0$ on RSU 1. Based on the hash value of $\text{Block}_0$, i.e. $\text{Hash}(\text{Block}_0)$, all nodes acquire the identity of the new committee leader by
\begin{equation}
\begin{gathered}
\text{ID}_L^1 \equiv \text{Hash}(\text{Block}_0) \Mod{M},
\end{gathered}
\end{equation}
where $M$ is the total number of RSUs in DBAFL. Assuming $\text{ID}_L^1 = 3$, RSU 3 is elected as the new committee leader at $\text{Block}_1$ and is placed in charge of aggregating and generating subsequent global models. The election frequency of the committee leader is the reciprocal of the number of blocks to wait before electing a new committee leader. Assume the frequency of the committee leader election is $1/10$, which means that the identity of the new committee leader is calculated in the same way as soon as $10$ blocks are appended to the blockchain. Take $\text{ID}_L^{11}=2$ as an example, RSU 2 is responsible for generating global models in $\text{Block}_{11}$ to $\text{Block}_{20}$.

Let the hash value of the latest block at time $t$ is $\text{Hash}(\text{Block}_{t-1})$ and the identity of RSU $m$ is $\text{ID}_m^t$. The probability that RSU $m$ becomes the leader at $t$ is denoted as $P_L(\text{ID}_m^t)$ and is calculated by
\begin{equation}
\begin{gathered}
P_L(\text{ID}_m^t) = \frac{P[\text{Hash}(\text{Block}_{t-1}) \Mod{M} \equiv \text{ID}_m]}{\sum_{j=1}^{M}P[\text{Hash}(\text{Block}_{t-1}) \Mod{M} \equiv \text{ID}_j]}.
\end{gathered}
\end{equation}
The duration of a specific RSU being a leader is limited to prevent it from creating forks on the blockchain and doing evil. Therefore, allowing all RSUs to have the same chance of becoming the leader is the best way to safeguard the system. Gini coefficient, a statistical measure of wealth inequality, is adopted to evaluate inequality in the probability of RSUs becoming a leader in DBAFL, which is calculated by
\begin{equation}
\begin{gathered}
G(t) = \frac{\sum^M_{m=1}\sum^M_{j=1}|P_L(\text{ID}_m^t) - P_L(\text{ID}_j^t)|}{2\sum^M_{m=1}\sum^M_{j=1}P_L(\text{ID}_j^t)}.
\end{gathered}
\end{equation}
It is empirically proved that when adopting SHA-256, a widely used hash function, the non-randomness percentage of the hash output is $31.25\%$~\cite{al2019randomness}. Accordingly, it is trivial to know that $G(t)$ is less than $0.3125$, which is close to $0$ and demonstrates that the probability of any RSUs becoming a leader in DBAFL is sufficiently even. With more hash functions are developed, adopting a more random hash function increases the equality level in the probability of RSUs becoming a leader and subsequently improves the security level of DBAFL.

The original models are stored in the distributed database on RSUs for buses to download. The hash values of the models are stored in the blocks and synchronized to all nodes for the purpose of validating the original model. If the downloaded local model is modified and inconsistent with the hash value stored on the blockchain, the committee leader would ignore it and subsequent local models from that node until a new committee leader is elected. Besides, the procedure of the global model aggregation is also verifiable by all committee members, preventing the committee leader from doing evil. 

\begin{table}[htpb]
    \rowcolors{2}{gray!20}{white}
    \renewcommand{\arraystretch}{1.3}
    \caption{The Comparison of Consensus Algorithms}
    \label{table:consensus-comparison}
    \centering
    \begin{tabular}{c|>{\centering\arraybackslash}p{0.14\linewidth}|>{\centering\arraybackslash}p{0.14\linewidth}|c|>{\centering\arraybackslash}p{0.15\linewidth}}
    \rowcolor{gray!50}
    \hline
    & \textbf{Energy Efficiency} & \textbf{Decentra-lization} & \textbf{Scalability} & \textbf{Model Evaluation} \\
    \hline
PoW~\cite{nakamoto2008bitcoin}			& No  & High   & Strong & No  \\
PoS~\cite{saad2021pos}					& Yes & Low    & Strong & No  \\
DPoS~\cite{lu2020blockchain}			& Yes & Low    & Strong & No  \\
PBFT~\cite{li2020scalable}				& Yes & High   & Low    & No  \\
DAG~\cite{silvano2020iota}				& Yes & High   & Strong & No  \\
Algorand~\cite{gilad2017algorand}		& Yes & High   & Strong & No  \\
The Proposed							& Yes & High   & Strong & Yes \\
    \hline
    \end{tabular}
\end{table}

Table~\ref{table:consensus-comparison} compares the proposed consensus algorithm with six representative ones. The qualitative comparison takes energy efficiency, decentralization, scalability, and the support of ML model evaluation into account, as all of these criteria are important for applying FL to public transportation systems. According to~\cite{bamakan2020survey}, PoW is energy-consuming, which is inappropriate for resource-limited vehicles. When evaluating from the aspects of governance, permission, and trust, PoS and DPoS have a relatively low decentralization level, which is more vulnerable to DDoS attacks than other consensus algorithms. By contrast, the proposed scheme is permissioned and trustable for buses and allows random RSUs to govern the transactions, which is more decentralized. PBFT has low scalability, which is not suitable for dynamic mobile networks. Although DAG and Algorand are energy-efficient, decentralized, and scalable, they do not support the model evaluation, as they are designed for cryptocurrency. Without model evaluation, the poisoned local models may degrade the performance of the global model. The proposed consensus algorithm is energy-efficient, decentralized, and scalable due to the hash-based committee leader election. The received local models are also evaluated by the committee leader using its local test dataset. By comparison, the proposed committee-based consensus algorithm is the most suitable one to support FL in the public transportation scenario.

\section{Model Analysis}
\label{sec:analysis}

In this section, DBAFL is theoretically analyzed from several aspects, including convergence, reliability, latency, mobility, and complexity.

\subsection{Convergence Analysis}

Assume there are $K$ nodes in DBAFL and $\mathcal{D}_k$ is the local data on node $k$. The number of samples on node $k$ is $n_k = |\mathcal{D}_k|$. $N$ is the total number of samples across $K$ nodes, which is calculated by $N = \sum^{K}_{k=1}|\mathcal{D}_k|$. Assume that $\forall k \neq k', \mathcal{D}_k \cap \mathcal{D}_{k'} = \varnothing$. The local empirical loss of node $k$ is:

\begin{equation}
\begin{gathered}
h_k(w_k) = \frac{1}{n_k} \sum_{i \in \mathcal{D}_k} \ell_i(w_k),
\end{gathered}
\end{equation}
where $\ell_i(w_k)$ is the corresponding loss function for data $i$ and $w_k$ is the local model parameter.
Considering the existence of $\epsilon$, the central objective function is calculated as:

\begin{equation}
\begin{gathered}
F(w) = \sum^{K}_{k=1} \frac{\epsilon_k}{K} h_k(w_k).
\label{eq:objective_function}
\end{gathered}
\end{equation}
where $w$ is the aggregated global model. The goal of Eq.~\ref{eq:objective_function} is to find a model that satisfies $w_* = \argmin_{w \in \mathbb{R}^d} F(w)$.

Following to~\cite{chen2020asynchronous}, suppose that $F(w)$ is $L$-smooth and $\mu$-strongly convex. The local functions $h_k(w)$ are $B$-locally dissimilar at $w$, then:

\begin{equation}
\begin{gathered}
\begin{aligned}
F(w^{t+1}) - F(w^t) & \leq 
-\nabla F(w^t)^\top \eta_k^t \frac{\epsilon_k'}{K} \nabla h_k(w^t) \\
& + \frac{L}{2}||\eta_k^t \frac{\epsilon_k'}{K}\nabla h_k(w^t)||^2,
\end{aligned}
\end{gathered}
\end{equation}
where $\eta_k=\frac{2\mu N'}{LB^2n_k'}$. $\mu$ is a non-negative value that satisfies $\mathbb{E} (\nabla h_k(w)) \leq ||\nabla F(w)||$. Since $\forall \epsilon_k>0$, $m_k = \eta_k^t \frac{\epsilon_k'}{K} > 0$. Assuming that $F(w)$ is bounded below, with the local bounded gradient dissimilarity defined in Chen \emph{et al.}~\cite{chen2020asynchronous}, it is trivial to know that,
\begin{equation}
\begin{gathered}
\mathbb{E}(F(w^{t+1})) - F(w^t) \leq -m_k(\mu - \frac{m_kLB^2}{2})||\nabla F(w^t)||^2
\end{gathered}
\end{equation}
is still monotonically increasing. In DBAFL, the accuracy of the initial global model must be greater than $1\%$. Therefore, $\epsilon_k < 100$. Assume there are at least $100$ total training samples among all nodes, $\epsilon_k' < K$. Therefore, $m_k = \eta_k^t \frac{\epsilon_k'}{K} < \eta_k^t$ and
\begin{equation}
\begin{gathered}
-m_k(\mu - \frac{m_kLB^2}{2}) < -\eta_k^t (\mu - \frac{\eta_k^t LB^2}{2}).
\end{gathered}
\end{equation}

So far, $\epsilon$ is already canceled out. As a result, the subsequent proofs are the same as the proofs of Theorem 1 and Theorem 2 in Chen \emph{et al.}~\cite{chen2020asynchronous}. Finally, it is proven that after $E$ epochs, DBAFL converges.

\subsection{Reliability Analysis}
\label{sec:reliability}

\subsubsection{Hash Values on Blockchain}
The benefits of uploading models to the blockchain instead of a centralized aggregation server are as follows: i) The consistency and reliability of global models are guaranteed since the data in the blockchain is immutable; ii) The training process becomes transparent and auditable, preventing nodes from doing evil; iii) Buses become trustable due to the existence of the consensus algorithm. 
Time efficiency is critical for smart public transportation systems, especially when training updated ML models for traffic condition prediction or driver assistance. Since the blockchain is resource-intensive, local nodes in DBAFL only upload the hash values of the models to the blockchain. The aforementioned benefits are preserved, as the model history is still traceable by the hash values in the blockchain and the original models are downloadable and verifiable by all local nodes. Furthermore, this mechanism greatly reduces the storage redundancy in the blockchain as well as the storage requirements for buses.

\subsubsection{Attack Resistance}
The design of DBAFL is primarily resistant to poisoning and DDoS attacks. When launching poisoning attacks, the attackers manipulate the parameters of local models and upload them to the global model, causing the accuracy of the global model to decrease~\cite{xu2021asynchronous}. However, the committee leader is able to identify the malicious local models by testing their accuracy locally. As a result, the committee leader will assign them relative low weight, which defends poisoning attacks to a certain extent. Since the committee-based consensus algorithm periodically elects a new leader based on the hash value of the latest block, the accuracy test result is ensured to be reliable and unbiased. In addition, DBAFL introduces a stricter defense strategy that discards local models with an accuracy below a specific threshold to further reduce the influence of malicious local models. DDoS attacks primarily aim to disrupt the centralized aggregation server in traditional FL schemes by flooding~\cite{doriguzzi2020lucid}. In DBAFL, the periodically changed committee leader replaces the centralized aggregation server in classic FL, reducing the probability of being targeted by traffic flooding in DDoS attacks.

\subsubsection{Leader Election}
A more frequent committee leader election leads to the increased reliability of DBAFL and the higher generality of the global model. However, it is almost impossible to guarantee efficiency if the committee leader is elected too frequently. Considering the mobile network is unstable, the network latency is likely to result in committee leader identities inconsistent and forks in the blockchain during block propagation. Aside from that, block size has an impact on efficiency as well. The decreased block size enables more frequent committee leader elections, which results in more blocks propagated in the blockchain and additional network overhead.

\subsubsection{Unstable Mobile Network}
Buses may fall offline unexpectedly in an unstable mobile network. As a comparison, RSUs equipping with backup for disaster recovery are more stable. Therefore, only RSUs are committee members and eligible to be the committee leader, ensuring a stable global model aggregation process. A bus that falls offline unexpectedly will not affect the training process on other nodes in DBAFL due to its asynchronous aggregation strategy. Storing models in InterPlanetary File System (IPFS) is a promising solution to further improve data reliability~\cite{henningsen2020mapping} and is left for future work.

\subsection{Latency Analysis}
\label{sec:latency}

According to~\cite{kim2019blockchained}, in a classic blockchain-based FL scheme, the latency of a training round is the sum of local training $T_{\text{local}}^{e}$, model uploading $T_{\text{up}}^{e}$, model aggregation $T_{\text{ag}}^{e}$, block generation $T_{\text{bg}}^{e}$, block propagation $T_{\text{bp}}^{e}$, and model downloading $T_{\text{dn}}^{e}$. Therefore, the latency of $e$-th epoch $T^{e}$ is given as
\begin{equation}
\begin{gathered}
T^{e} = T_{\text{local}}^{e} + T_{\text{up}}^{e} + T_{\text{ag}}^{e} + T_{\text{bg}}^{e} + T_{\text{bp}}^{e} + T_{\text{dn}}^{e}.
\end{gathered}
\end{equation}

Specifically, in the local training phase, nodes do not need communication with others. Since the asynchronous aggregation strategy does not wait for the local training on nodes before aggregation, the local training latency $T_{\text{local}}^{e}$ is ignored in DBAFL. As illustrated in Line 11, 12, and 17 of Algorithm~\ref{alg:async-fl}, the latency of model uploading phase is composed of three parts: local model uploading $T_{\text{up\_model}}^{e}$ from buses to RSUs, model hash uploading $T_{\text{up\_hash}}^{e}$ from buses to RSUs, and model synchronization $T_{\text{sync\_model}}^{e}$ among RSUs. Therefore, the latency of model uploading phase $T_{\text{up}}^{e}$ is expressed as
\begin{equation}
\begin{gathered}
\begin{aligned}
T_{\text{up}}^{e} & = T_{\text{up\_model}}^{e} + T_{\text{up\_hash}}^{e} + T_{\text{sync\_model}}^{e} \\
& = \frac{S_{w}}{\mathbb{B}_{M}\log_{2}(1+\gamma_{M})} + \frac{S_{\text{hash}}}{\mathbb{B}_{M}\log_{2}(1+\gamma_{M})} + \frac{S_{w}}{\mathbb{B}_{E}},
\end{aligned}
\end{gathered}
\end{equation}
where $S_{w}$ is the size of the model, $S_{\text{hash}}$ is the size of the model hash, $\mathbb{B}_{E}$ and $\mathbb{B}_{M}$ are the bandwidth allocations of the Ethernet network and the mobile network, respectively, and $\gamma_{M}$ is the received signal-to-noise ratio (SNR) of the devices in the mobile network. As RSUs are connected with each other through a high-speed Ethernet, there is no SNR considered in $T_{\text{sync\_model}}^{e}$.

After aggregation, the hash value of the global model is uploaded to the blockchain while the original global model is synchronized among RSUs, as shown in Line 22 and 23 of Algorithm~\ref{alg:async-fl}. Assuming the time of processing aggregation is negligible compared with the communication delays, the latency of model aggregation phase $T_{\text{ag}}^{e}$ is calculated as
\begin{equation}
\begin{gathered}
T_{\text{ag}}^{e} = T_{\text{up\_hash}}^{e} + T_{\text{sync\_model}}^{e} = \frac{S_{\text{hash}}}{\mathbb{B}_{M}\log_{2}(1+\gamma_{M})} + \frac{S_{w}}{\mathbb{B}_{E}}.
\end{gathered}
\end{equation}

In order to improve efficiency, the committee-based consensus algorithm in DBAFL does not involve mining or communicating during the block generation process, as demonstrated in Section~\ref{sec:consensus_algorithm}. Therefore, the latency of generating blocks $T_{\text{bg}}^{e}$ only involves a small amount of time spent computing hash values and is considered negligible in comparison to the communication delays. Thereafter, the newly generated block is propagated throughout the network with latency
\begin{equation}
\begin{gathered}
T_{\text{bp}}^{e} =  \frac{S_{\text{block}}}{\mathbb{B}_{M}\log_{2}(1+\gamma_{M})},
\end{gathered}
\end{equation}
where $S_{\text{block}}$ is the size of the block. Finally, as shown in Line 9 of Algorithm~\ref{alg:async-fl}, buses download the new global model from the nearby RSU with latency
\begin{equation}
\begin{gathered}
T_{\text{dn}}^{e} = \frac{S_w}{\mathbb{B}_{M}\log_{2}(1+\gamma_{M})}.
\end{gathered}
\end{equation}

Moreover, compared with an asynchronous federated learning scheme, the additional communication latency brought by the blockchain $T_{\text{bc}}^{e}$ is
\begin{equation}
\begin{gathered}
T_{\text{bc}}^{e} = 2 T_{\text{up\_hash}}^{e} + 2 T_{\text{sync\_model}}^{e} + T_{\text{bp}}^{e},
\end{gathered}
\end{equation}
due to the requirements of uploading hash values of global and local models, synchronizing local and global models among RSUs, and releasing the new block. 
Since the hash size is much smaller than the model size, $S_{\text{hash}}$ and $S_{\text{block}}$ are much smaller than $S_{w}$. Thus, $T_{\text{up\_hash}}^{e} + T_{\text{bp}}^{e} \ll T_{\text{up}}^{e} + T_{\text{dn}}^{e}$.
Moreover, by increasing the bandwidth among RSUs $\mathbb{B}_{E}$, $T_{\text{sync\_model}}^{e}$ is easy to be reduced to very small. As a result, $T_{\text{bc}}^{e}$ is smaller than $T^{e}$ and acceptable in public transportation scenarios.

\subsection{Mobility Analysis}
\label{sec:mobility}

Assume buses are traveling on a road in a built-up area, where the 5G network coverage of an RSU is 300 meters and the vehicle speed limit is 60 km/h~\cite{shafi20175g}. By calculating, the connection of a running bus to an RSU lasts 18 seconds at most. Due to the device heterogeneity, it is hard to determine how long it will take a bus to finish training a local model. In DBAFL, the training of local models is independent for each bus due to the asynchronous aggregation strategy, which means that no network connection or waiting for others is required during the training process. After training, buses need to upload the local model to the nearby RSU and the hash value to the blockchain before downloading a new global model from the nearby RSU. As evaluated in Section~\ref{sec:rq2_time_costs}, the communication time cost in each training round of DBAFL is always less than 5 seconds, which is much less than the limitation of 18 seconds. After receiving the local model, the committee leader performs aggregation locally and generates a new global model, which is shared with other RSUs without effects from the mobility of buses. Besides, the leader election among committee members (RSUs) is also not affected by the mobility of buses. Therefore, considering the mobility of buses, DBAFL is still feasible in smart public transportation systems.

\subsection{Complexity Analysis}
\label{sec:complexity}
As the collected data changes over time, models are trained on a regular basis. Assuming the newly arrived data size on a bus is $n$, the computational complexity of training on the bus is $\mathcal{O}(n)$. Since each bus has to perform training for $E$ epochs, the computational complexity of DBAFL on each bus is $\mathcal{O}(nE)$. Considering DBAFL enables buses to train parallelly without waiting for models from others, the overall computational complexity of DBAFL is $\mathcal{O}(nE)$, which is acceptable.

Compared with classic FL, DBAFL has an additional communication complexity of uploading hash of models to the blockchain and electing new committee leaders. As the hash values are small enough without effects of the model size, it is easy to be packed into blocks and broadcast to nodes through the P2P protocol. Since the identity of the new committee leader is determined by the hash value of the specific block, no more network communication is required after the block is broadcasted to local nodes in IoV networks. Therefore, the communication complexity for each training round is $\mathcal{O}(\log{}K)$, where $K$ is the number of nodes in DBAFL. Besides, the blockchain could be deployed purely on RSUs to reduce the communication complexity of buses at the cost of certain security and credibility. In that situation, the communication complexity for each training round is $\mathcal{O}(\log{}M)$, where $M$ ($M \ll K$) is the number of RSUs.

\section{System Evaluation}
\label{sec:system_evaluation}

In this section, experiments are conducted from three aspects, including learning performance, efficiency, and reliability, to evaluate DBAFL on IoV networks and answer the following research questions:

\begin{itemize}
    \item \textbf{RQ1:} How well does DBAFL improve learning performance compared with state-of-the-art schemes?
    \item \textbf{RQ2:} Is there any advantage of DBAFL in efficiency compared with state-of-the-art schemes?
    \item \textbf{RQ3:} Is DBAFL reliable enough to resist poisoning and DDoS attacks?
\end{itemize}

\subsection{Experiment Setup}

Five virtual machines (VM) and four Raspberry PI B4 devices are set up as the experiment environment. Each VM has 8 CPU cores and 8GB RAM to simulate an RSU. Each Raspberry PI B4 has 4 CPU cores and 8GB RAM to simulate a vehicle with limited computing resources. In terms of software configuration, the Ubuntu 20.10 operating system is deployed on all nodes. ML models are trained with PyTorch v1.8.1 based on Python 3.8. The smart contract is developed on Hyperledger Fabric v2.3.0, an open-source blockchain framework, to orchestrate model training and aggregation on nodes and accept model hash values. Specifically, the smart contract is in charge of reading and writing the uploaded hash of models on the blockchain. Besides, whenever a new global model is generated (the hash of the new global model is received), the smart contract drives each node to download the latest global model from the committee leader for the next round of training. A RESTful service is developed on Express.js v4.17.1, allowing local nodes to upload or download the hash values of models from the blockchain. The implementation details are available at \url{https://github.com/xuchenhao001/AFL}.

The default parameter settings for experiments are shown in Table~\ref{table:blockchain-parameter-setting}. The number of nodes on a road is assumed to be five by default, including an RSU and four buses traveling under the network coverage of the RSU. In experiments, the number of buses and RSUs ranges from one to four. Similar to~\cite{mcmahan2017communication,xu2021scei,xu2021bafl}, the local data size $B$ is $1500$, the number of local epochs $E$ is $50$, and the learning rate $\eta$ is $0.01$. To examine the effectiveness of the dynamic setting strategy, the value of $\epsilon$ is set to static, including $0.5$, $1.0$, and $1.5$, to reveal the impact of under-, equal-, and over-weighted stale local models. According to the default parameter settings in Hyperledger Fabric~\cite{androulaki2018hyperledger}, a new block is generated when any of the following conditions is reached: the waiting time for the next invoke reaches $2$ seconds, the number of invokes reaches $10$, or the block size reaches $10$ MB.

\begin{table}
    \rowcolors{2}{gray!20}{white}
    \renewcommand{\arraystretch}{1.3}
    \caption{The Experiment Parameter Settings}
    \label{table:blockchain-parameter-setting}
    \centering
    \begin{tabular}{c|c}
    \rowcolor{gray!50}
    \hline
    \textbf{Parameter} & \textbf{Value}\\
    \hline
    The number of nodes $K$
    & $5$\\
    The local data size $B$
    & $1500$\\
    The number of epochs $E$
    & $50$\\
    The learning rate $\eta$
    & $0.01$\\
    The static scaling factor $\epsilon$
    & \{$0.5$, $1.0$, $1.5$\}\\
    The time to wait before creating a block
    & $2$s\\
    The maximum number of messages in a block
    & $10$\\
    The maximum bytes of messages in a block
    & $10$MB\\
    \hline
    \end{tabular}
\end{table}

The ML models are trained on two benchmark datasets (i.e. CIFAR-10~\cite{xu2021bafl} and FMNIST~\cite{xiao2017fashion}) and one real-world dataset (i.e. LOOP~\cite{cui2019traffic}). Specifically, the LOOP dataset contains the speed information collected by the inductive loop detectors deployed on freeways in the Seattle area at intervals of $5$ minutes. The ML models include MLP, CNN, and LSTM. 

To evaluate DBAFL, several state-of-the-art schemes are included in the scope for comparison.
\begin{enumerate}
    \item \textbf{BSFL}: The synchronized version of DBAFL without the dynamic scaling factor. Blockchain is included.
    \item \textbf{ASOFED}: An AFL scheme with static decay coefficient balancing the previous and current gradients~\cite{chen2020asynchronous}. Blockchain is not included.
    \item \textbf{BDFL}: An AFL scheme with all models saved in the blockchain~\cite{chen2021bdfl}.
    \item \textbf{APFL}: A semi-AFL scheme with reduced the communication frequency~\cite{deng2020adaptive}. Blockchain is not included.
    \item \textbf{FedAVG}: The traditional synchronous FL scheme~\cite{mcmahan2017communication}. Blockchain is not included.
    \item \textbf{Local}: The traditional local training scheme that each node trains the ML model on local data without communication. Blockchain is not included.
\end{enumerate}

\begin{figure*}[htb]
  \begin{center}
    \subfigure[CNN model on CIFAR-10]{
      \includegraphics[width=0.32\linewidth]{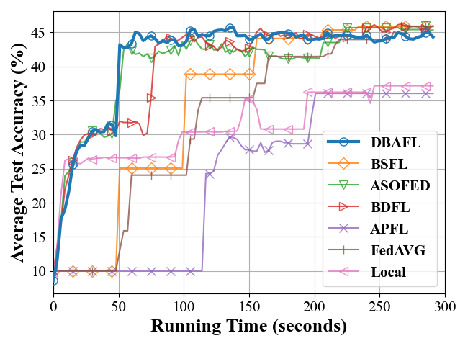}}
    \subfigure[CNN model on FMNIST]{
      \includegraphics[width=0.32\linewidth]{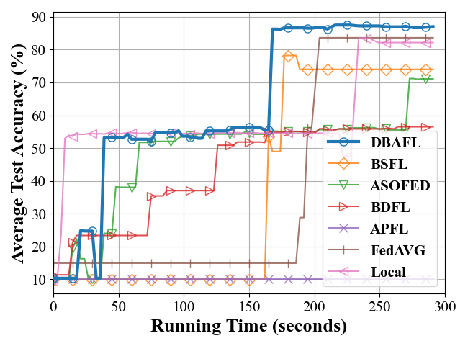}}
    \subfigure[MLP model on FMNIST]{
      \includegraphics[width=0.32\linewidth]{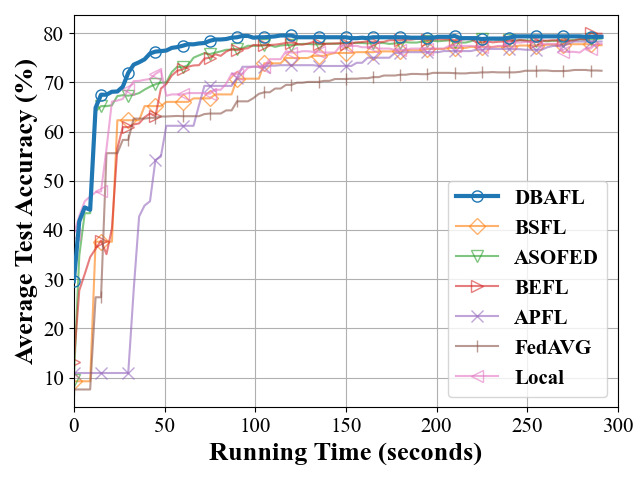}}
  \end{center}
  \caption{Compare the average test accuracy of models in DBAFL with that of models in other schemes. There are five nodes in the network, two of which are vehicles while others are RSUs.}
  \label{fig:compare_performance_others}
\end{figure*}

\begin{figure*}[htb]
  \begin{center}
    \subfigure[One Vehicle]{
      \includegraphics[width=0.238\linewidth]{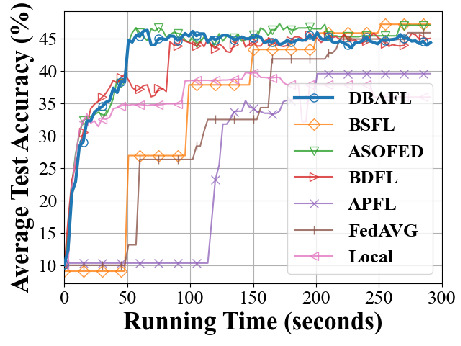}}
    \subfigure[Two Vehicles]{
      \includegraphics[width=0.238\linewidth]{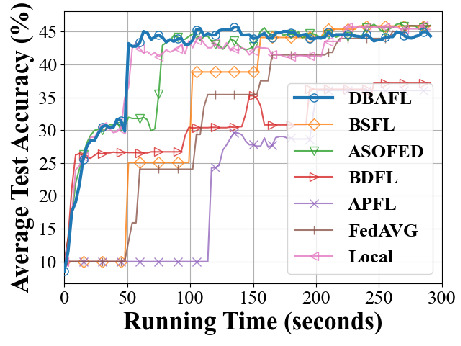}}
    \subfigure[Three Vehicles]{
      \includegraphics[width=0.238\linewidth]{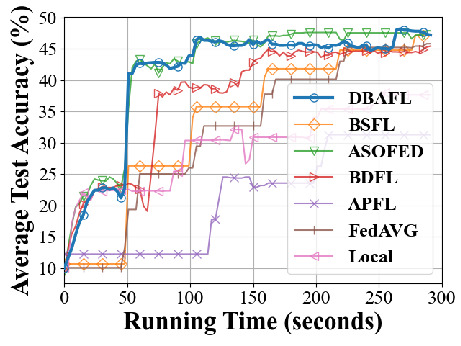}}
    \subfigure[Four Vehicles]{
      \includegraphics[width=0.238\linewidth]{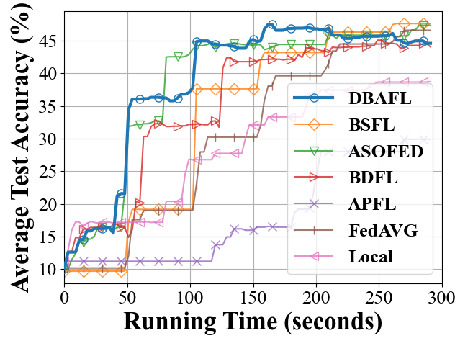}}

    \subfigure[One Vehicle]{
      \includegraphics[width=0.238\linewidth]{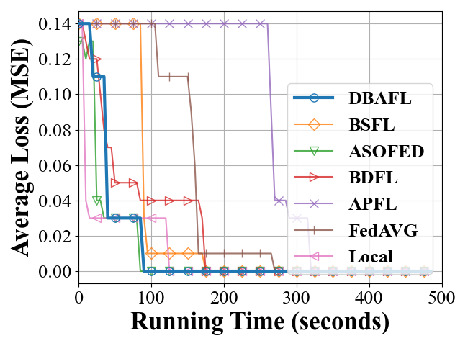}}
    \subfigure[Two Vehicles]{
      \includegraphics[width=0.238\linewidth]{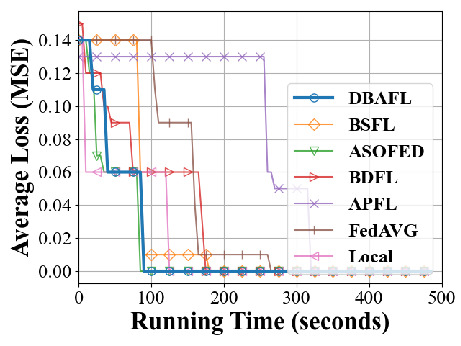}}
    \subfigure[Three Vehicles]{
      \includegraphics[width=0.238\linewidth]{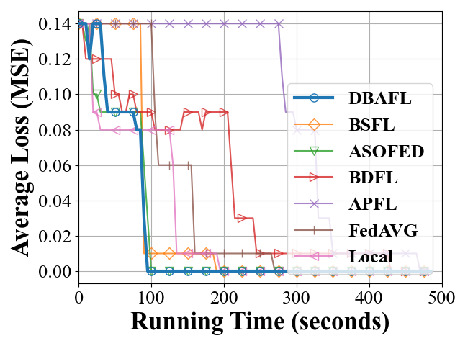}}
    \subfigure[Four Vehicles]{
      \includegraphics[width=0.238\linewidth]{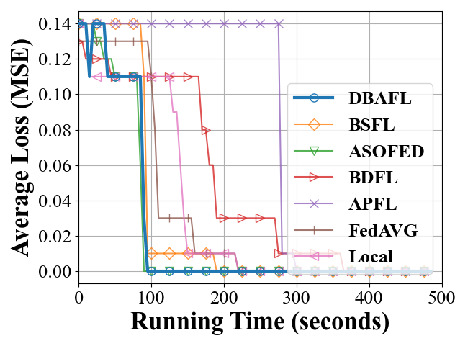}}
  \end{center}
  \caption{Compare the average test accuracy when different numbers of vehicles participate in training. There are five nodes in the network. Top row: The average test accuracy when train CNN on CIFAR-10. Bottom row: The average loss evaluated by mean squared error when train LSTM on LOOP.}
  \label{fig:compare_performance_iot}
\end{figure*}

\begin{figure*}[htb]
  \begin{center}
    \subfigure[One Big Node]{
      \includegraphics[width=0.238\linewidth]{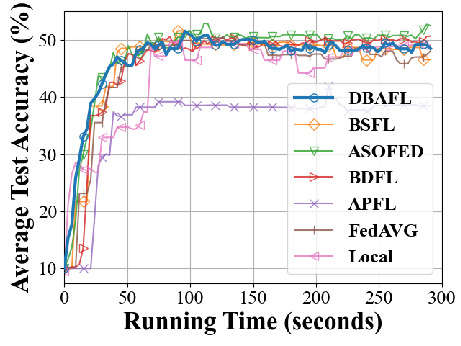}}
    \subfigure[Two Big Nodes]{
      \includegraphics[width=0.238\linewidth]{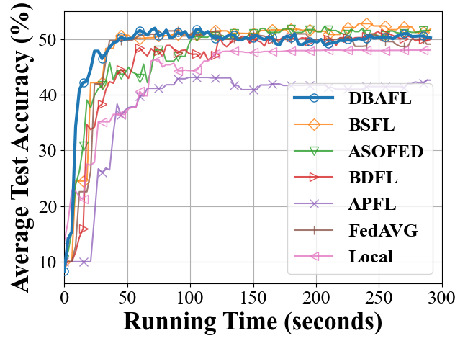}}
    \subfigure[Three Big Nodes]{
      \includegraphics[width=0.238\linewidth]{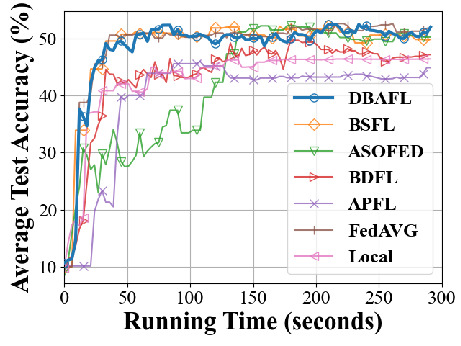}}
    \subfigure[Four Big Nodes]{
      \includegraphics[width=0.238\linewidth]{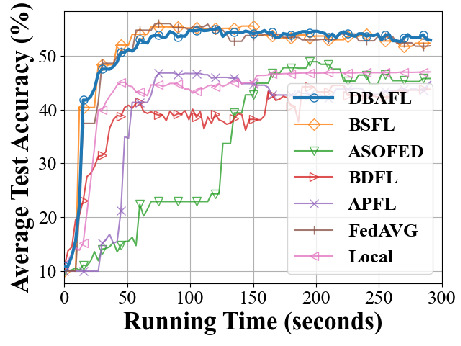}}

    \subfigure[One Big Node]{
      \includegraphics[width=0.238\linewidth]{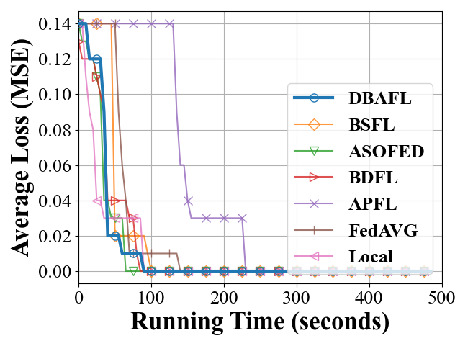}}
    \subfigure[Two Big Nodes]{
      \includegraphics[width=0.238\linewidth]{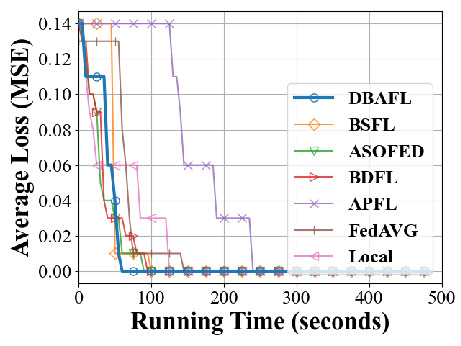}}
    \subfigure[Three Big Nodes]{
      \includegraphics[width=0.238\linewidth]{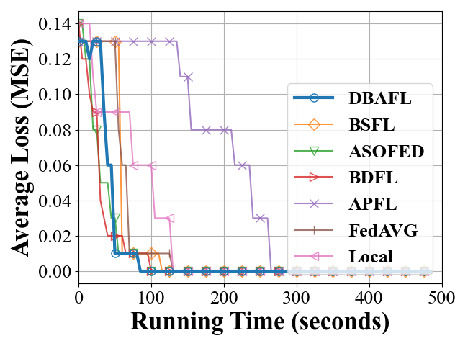}}
    \subfigure[Four Big Nodes]{
      \includegraphics[width=0.238\linewidth]{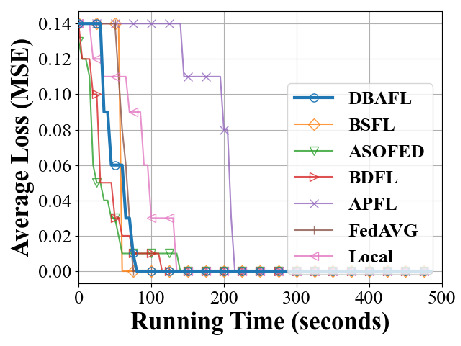}}
  \end{center}
  \caption{Compare the average test accuracy when different numbers of big nodes participate in training. There are five nodes in the network. Top row: The average test accuracy when train CNN on CIFAR-10. Bottom row: The average loss evaluated by mean squared error when train LSTM on LOOP.}
  \label{fig:compare_performance_dis}
\end{figure*}

\begin{figure*}[htb]
  \begin{center}
    \subfigure[CNN model on CIFAR-10]{
      \includegraphics[width=0.32\linewidth]{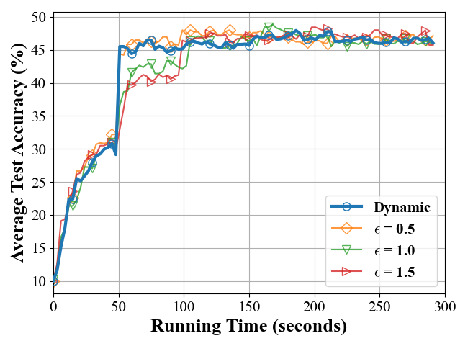}}
    \subfigure[CNN model on FMNIST]{
      \includegraphics[width=0.32\linewidth]{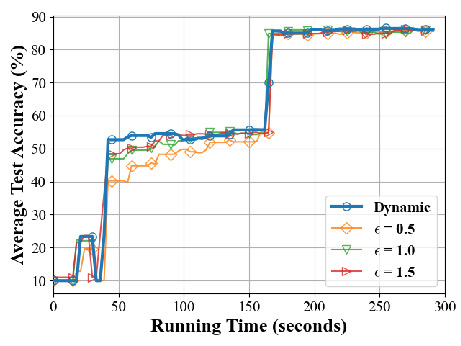}}
    \subfigure[MLP model on FMNIST]{
      \includegraphics[width=0.32\linewidth]{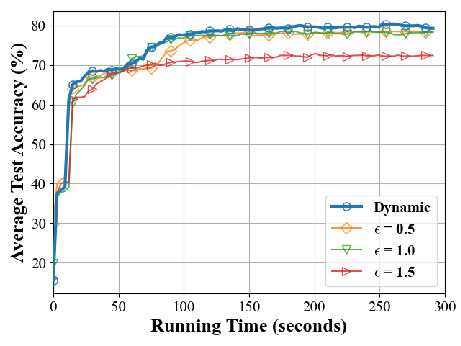}}
  \end{center}
  \caption{Compare the average test accuracy under dynamic scaling factor setting with that under static scaling factor settings. There are five nodes in the network, two of which are vehicles while others are RSUs.}
  \label{fig:compare_performance_static}
\end{figure*}

\begin{figure*}[htb]
  \begin{center}
    \subfigure[CNN model on CIFAR-10]{
      \includegraphics[width=0.238\linewidth]{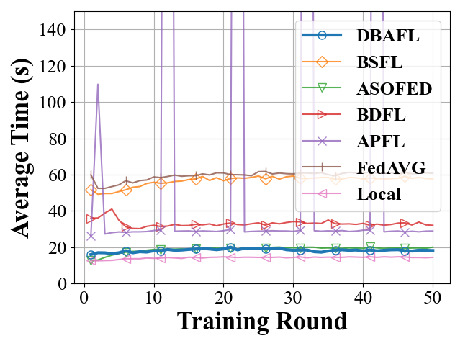}}
    \subfigure[CNN model on FMNIST]{
      \includegraphics[width=0.238\linewidth]{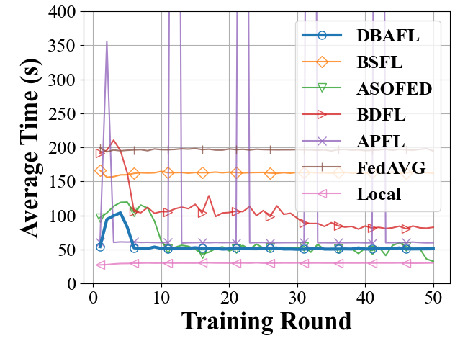}}
    \subfigure[MLP model on FMNIST]{
      \includegraphics[width=0.238\linewidth]{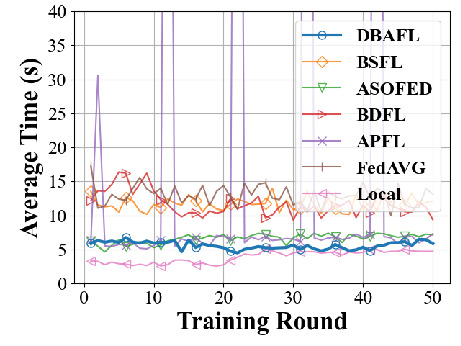}}
    \subfigure[LSTM model on LOOP]{
      \includegraphics[width=0.238\linewidth]{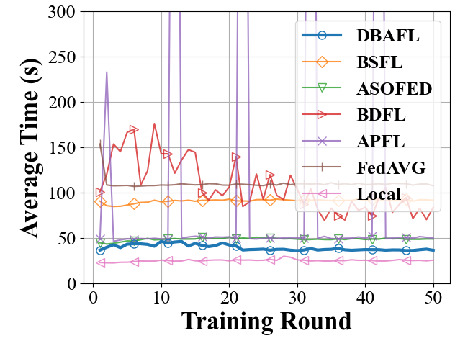}}
  \end{center}
  \caption{Compare the average overall time cost in each training round of DBAFL with other schemes. There are five nodes in the network, two of which are vehicles while others are RSUs.}
  \label{fig:compare_time_overall}
\end{figure*}

\begin{figure*}[htb]
  \begin{center}
    \subfigure[CNN model on CIFAR-10]{
      \includegraphics[width=0.238\linewidth]{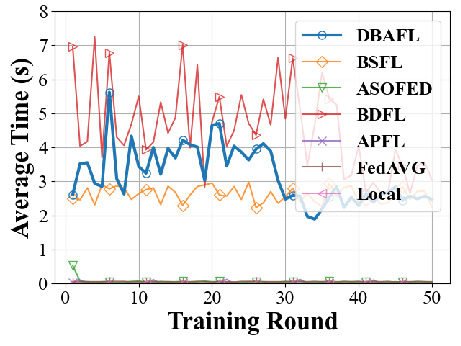}}
    \subfigure[CNN model on FMNIST]{
      \includegraphics[width=0.238\linewidth]{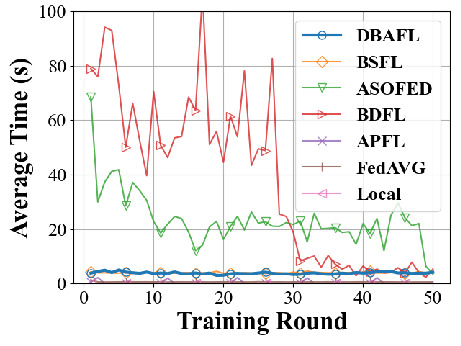}}
    \subfigure[MLP model on FMNIST]{
      \includegraphics[width=0.238\linewidth]{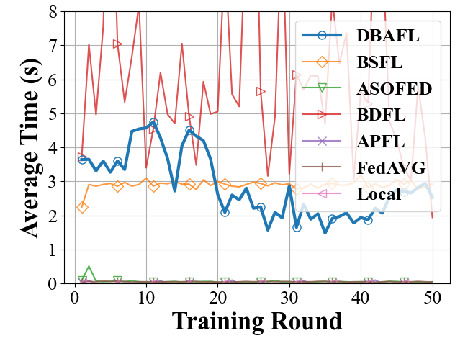}}
    \subfigure[LSTM model on LOOP]{
      \includegraphics[width=0.238\linewidth]{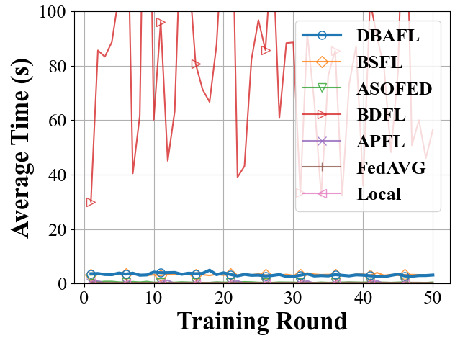}}
  \end{center}
  \caption{Compare the average communication time cost in each training round of DBAFL with other schemes. There are five nodes in the network, two of which are vehicles while others are RSUs.}
  \label{fig:compare_time_communication}
\end{figure*}

\begin{figure}[htb]
    \centering
    \includegraphics[width=\linewidth]{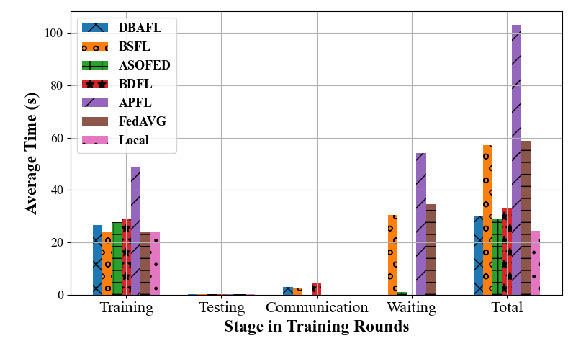}
    \caption{Compare the average time cost of four different stages, including training, test, communication, and waiting, in the training round when training the CNN model on the CIFAR-10 dataset. Training: Nodes train their local models; Testing: Nodes test the accuracy of the local and global models; Communication: Nodes upload and download local and global models; Waiting: Nodes wait for the global model to be aggregated. Total: The sum of the average time cost of four stages. There are five nodes in the network, two of which are vehicles while others are RSUs.}
    \label{fig:compare_time_cost_type}
\end{figure}

\begin{figure}[htb]
    \centering
    \includegraphics[width=0.8\linewidth]{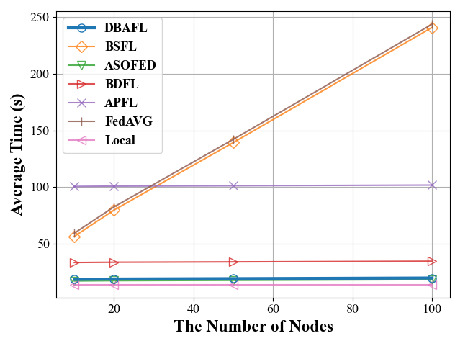}
    \caption{Compare the average time cost when training the CNN model on the CIFAR-10 dataset with different numbers of nodes in the network. All nodes are with the same computing resources as RSUs.}
    \label{fig:compare_time_cost_node}
\end{figure}

\begin{figure*}[htb]
  \begin{center}
    \subfigure[Classic AFL]{
      \includegraphics[width=0.238\linewidth]{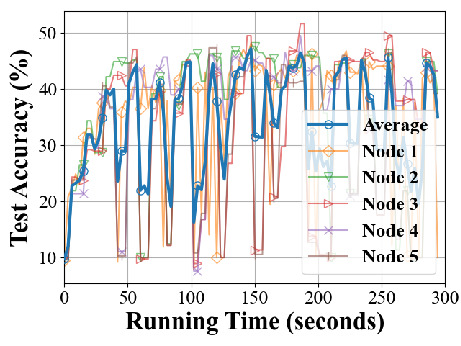}}
    \subfigure[DBAFL without defense]{
      \includegraphics[width=0.238\linewidth]{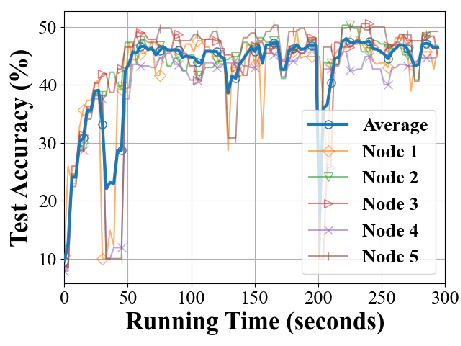}}
    \subfigure[DBAFL with 80\% defense]{
      \includegraphics[width=0.238\linewidth]{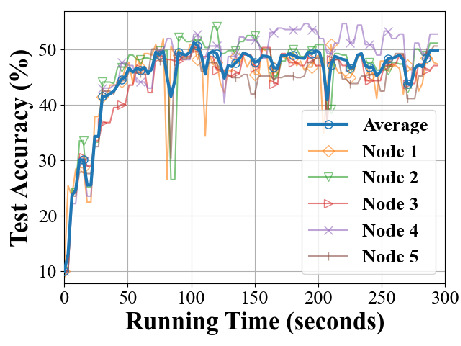}}
    \subfigure[DBAFL with 90\% defense]{
      \includegraphics[width=0.238\linewidth]{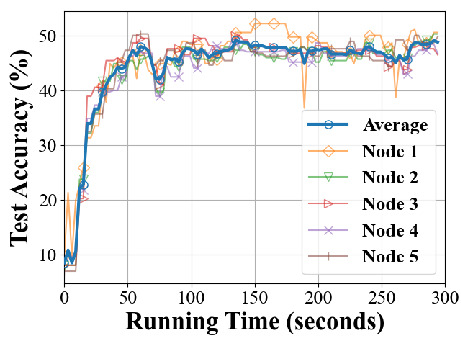}}
  \end{center}
  \caption{When Node 5 launches poisoning attacks, compare the average test accuracy of models in DBAFL at various degrees of defense thresholds with that of models in the classic AFL scheme without any defense.}
  \label{fig:compare_attack_poisoning}
\end{figure*}

\begin{figure}[htb]
    \centering
    \subfigure[Classic AFL]{
      \includegraphics[width=0.48\linewidth]{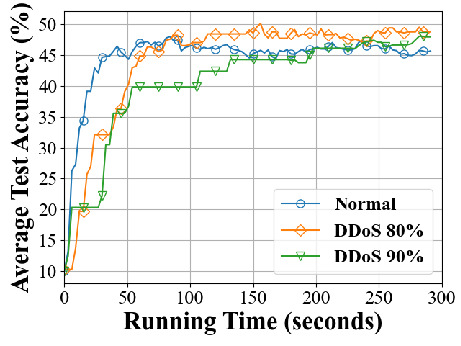}}
    \subfigure[DBAFL]{
      \includegraphics[width=0.48\linewidth]{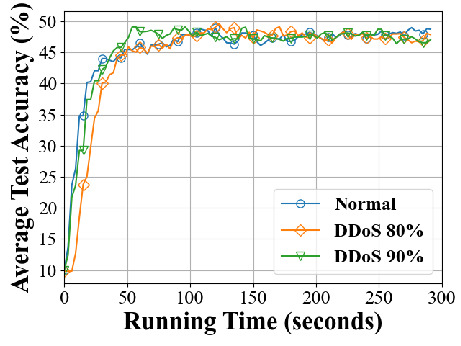}}
    \caption{When suffering different degrees of DDoS attacks, compare the average test accuracy of models in DBAFL with that of models in the classic AFL scheme.}
    \label{fig:compare_attack_ddos}
\end{figure}

Average test accuracy is calculated by averaging the accuracy of the local models on all nodes in a fixed-time interval. The average loss is calculated in the same way based on the mean square error (MSE). The average time is calculated by averaging the time cost of all nodes in each training round. The average time per iteration is calculated by averaging the time cost of all nodes across all training rounds. 

To answer \emph{RQ1}, the average test accuracy is compared with state-of-the-art schemes when different models, datasets, and IoV network settings are applied. Specifically, there are two aspects of IoV network settings: (1) using different numbers of IoT devices and VMs to mimic the differences in computing resources among vehicles and RSUs; (2) distributing different sizes of local data on VMs to mimic the disparity of data collected among nodes. The node with a large volume of data is the big node, whose local data size is set to $5000$. For clear comparison, the local data size for a small node is set to $500$. In addition, to validate the effectiveness of the dynamic setting, the average test accuracy with the scaling factor adopted under three static settings, including $0.5$, $1.0$, and $1.5$, is compared with that under the dynamic setting.

To answer \emph{RQ2}, the average overall time cost in each training round is compared when adopting different models and datasets in experiments. To further investigate the communication overhead brought by the blockchain, the average communication time cost in each training round is also compared. When training the CNN model on the CIFAR-10 dataset, the average time costs on different stages and in different scales of networks are compared.

To answer \emph{RQ3}, Node 5 randomly adjusts the parameters in local models before sending them to the committee leader to simulate poisoning attacks. The test accuracy of DBAFL with different levels of defense strategies is compared with that of the classic AFL scheme without any defense strategy. Moreover, the average test accuracy of DBAFL and that of the classic AFL scheme are compared at different levels of DDoS attacks, especially when $80\%$ or $90\%$ of the total traffic is the DDoS attack traffic.

\subsection{Results Analysis}

\subsubsection{RQ1. Convergence Speed and Model Accuracy}

As shown in Fig.~\ref{fig:compare_performance_others}, initially, the convergence speed of DBAFL is a little lower than that of the local training scheme but higher than the other ones. Subsequently, relatively early in the cycle of the tests, the average test accuracy of DBAFL steps up to an optimal level and converges faster than all other schemes. The step-up is due to the contribution of stale local models from vehicles. When training the CNN model, the step-up happens earlier on CIFAR-10 (at around the $50$th second) than that on FMNIST (at around the $160$th second), revealing that FMNIST is more complex and harder for vehicles to train. In addition, on the FMNIST dataset, DBAFL converges faster when the trained model is MLP, rather than CNN, because the MLP model is much simpler than the CNN model. Finally, the average test accuracy of DBAFL when training CNN and MLP models on two different datasets is always optimal compared with that of other schemes, which reveals the stable learning performance of DBAFL.

When training the CNN model on the CIFAR-10 dataset, as shown in Fig.~\ref{fig:compare_performance_iot} (a) to (d), the advantage of DBAFL in terms of convergence speed is more obvious compared with BSFL, BDFL, APFL, and FedAVG, under the circumstance of fewer vehicles. For example, the convergence time of DBAFL is around $150$ seconds earlier (note the $50$th second and the $200$th second marks) than that of BSFL and FedAVG when only one vehicle is in the network, while the convergence time of DBAFL is around $50$ seconds earlier (note the $150$th second and the $200$th second marks) than that of BSFL and FedAVG when four vehicles are in the network. The reason is that more nodes with rich computing resources involved in the network contribute more local models at the early stage, resulting in higher convergence speeds. When comparing the average test accuracy of models, DBAFL performs more stable (at around $45\%$) than the local training scheme (from around $40\%$ down to around $20\%$) as the number of vehicles increases. This is because a lagging node with limited computing resources is unable to learn information from the high-performance nodes in the local training scheme. As a result, more lagging nodes in the network lead to lower average test accuracy among all nodes.
When training the LSTM model on the LOOP dataset, as shown in Fig.~\ref{fig:compare_performance_iot} (e) to (h), the convergence speed of DBAFL also decreases (as expected) with the increase of vehicles in the network, although the final convergence time is almost identical. The reason is that the LSTM model fits the LOOP dataset easily and convergences after the first round of aggregation, while the finish time of the first round of training on vehicles is almost the same (at around the $100$th second mark).

To assess the impact of local data size on average test accuracy, big nodes (nodes with a lot of data) and small nodes (nodes with a small amount of data) have $5000$ and $500$ training samples, respectively. As shown in Fig.~\ref{fig:compare_performance_dis} (a) to (d), DBAFL is able to achieve the optimal model accuracy compared with the state-of-the-art schemes. With more big nodes in the network, ASOFED and BDFL have decreased model accuracy and slowed convergence speed. This is because too much training data on big nodes slows down their training process, while the local models from small nodes are not weighted appropriately during the aggregation.
When training the LSTM model on the LOOP dataset, as shown in Fig.~\ref{fig:compare_performance_dis} (e) to (h), the convergence speed of DBAFL is barely affected by any increase in the number of big nodes in the network, which is determined by the fastest node in the network. Nevertheless, compared with other schemes, DBAFL is always the fastest to converge.

Compared with static settings, the dynamic setting strategy allows DBAFL to assign an optimal scaling factor during the training process, resulting in the best convergence speed and model accuracy in all situations, as shown in Fig.~\ref{fig:compare_performance_static}. In addition, the ideal static scaling factor setting, which varies when training various models on different datasets, follows no obvious pattern. Take the CNN model as an example: the ideal static setting for the scaling factor when training on CIFAR-10 is $0.5$, whereas it is $1.5$ when training on FMINIST. Because the learning process is random, all nodes have the same probability of discovering an appropriate learning direction, regardless of computing resources or local data size. When the fast nodes discover the best learning direction first, it is desirable to set the scaling factor to a value less than $1.0$ in order to reduce the impact of stale local models. On the contrary, it is preferential to set the scaling factor to a value greater than $1.0$ to amplify the impact of remarkable local models when the fast nodes initially identify the worst learning direction.

\begin{framed}
\noindent Result 1: DBAFL has a superior convergence speed and optimal model accuracy compared with state-of-the-art schemes.
\end{framed}

\subsubsection{RQ2. Time Costs}
\label{sec:rq2_time_costs}

As shown in Fig.~\ref{fig:compare_time_overall}, DBAFL has the lowest overall time cost in each training round compared with state-of-the-art schemes in all situations. Especially, when training the CNN model on the FMNIST dataset, the average overall time cost of DBAFL in each training round (around $50$ seconds) is $150$ seconds less than that of FedAVG (around $200$ seconds). This is because the asynchronous aggregation strategy shortens the waiting time before aggregation. As a result, the advantage of DBAFL in terms of time cost in each training round becomes ever more apparent if the training process is more time-consuming. In addition, the periodic spike in average overall time cost in APFL is mainly caused by the waiting for the lagging nodes in every 10 rounds of training.

The average communication time cost in each training round is in line with the previous analysis, as shown in Fig.~\ref{fig:compare_time_communication}. The average communication cost in each training round of DBAFL is higher than that of FedAVG and APFL, and comparable to that of BSFL. This is caused by the additional time cost in the underlying blockchain architecture, including consensus and block propagation. However, it is obvious that DBAFL has a lower average communication time cost than BDFL, since hash values instead of the original models are uploaded to the blockchain. Moreover, compared with the average overall time cost in each training round of classic FL (FedAVG), the additional communication time cost incurred due to blockchain is mostly negligible, especially when training the CNN model on the FMNIST dataset (at $200$ seconds compared with $4$ seconds). Since the additional communication time cost brought by blockchain is $3$ seconds with little variation, the impact of blockchain becomes less as the model training task becomes more complex.

When training the CNN model on the CIFAR-10 dataset, the average time cost at different stages among all training rounds and all nodes is summarized in Fig.~\ref{fig:compare_time_cost_type}. In terms of the training stage, DBAFL has a little higher average time cost than BSFL, FedAVG, and Local Training (around $2.5$ seconds higher), which is due to the more frequent aggregation requests sent to the committee leader under the asynchronous aggregation strategy. APFL has the highest average time cost on training (almost twice the time than for the other schemes) due to two training procedures in each training round. In terms of the test stage, it is obvious that all schemes spend very little time (less than $0.3$ seconds), implying that the additional accuracy-test stage in DBAFL has a negligible impact on the efficiency of AFL. In terms of the communication stage, the average time cost of DBAFL, BSFL, and BDFL is slightly higher than that of FedAVG, APFL, and Local Training due to the consensus process of the blockchain. However, DBAFL has a minimal time cost in waiting for other nodes, which is similar to ASOFED, BDFL, and Local Training, due to its asynchronous aggregation strategy. On the other hand, BSFL, APFL, and FedAVG waste nearly half of the time in a round waiting instead of training or communicating, resulting in lower training efficiency when compared with DBAFL. After summing the average time cost of training, test, communication, and waiting stages, the total round time of DBAFL is only $0.97$ seconds longer than that of ASOFED. This shows that the proposed scheme effectively mitigates the effects of the blockchain.

When increasing the network scale from $10$ to $100$, the average time cost of a training round is demonstrated in Fig.~\ref{fig:compare_time_cost_node}. As the number of nodes in the network increases, high-performance nodes have to wait for more lagging nodes in each training round when adopting synchronous aggregation strategies, implying a longer waiting time. As a result, the schemes utilizing asynchronous aggregation strategies (DBAFL, ASOFED, BDFL, and APFL) have better scalability than those adopting synchronous aggregation strategies (BSFL and FedAVG). Besides, even with the blockchain incorporated, DBAFL achieves the same scalability as the pure AFL scheme ASOFED and has higher scalability than other blockchain-based schemes, due to the adoption of the proposed efficient consensus algorithm.

\begin{framed}
\noindent Result 2: DBAFL spends the shortest time in each training round while making full use of the computing resources on each node without time wasted waiting for other nodes.
\end{framed}

\subsubsection{RQ3. Attack Resistance}

As shown in Fig.~\ref{fig:compare_attack_poisoning} (a) and (b), the classic AFL scheme has difficulty converging under poisoning attacks, while DBAFL is resistant to poisoning attacks initially to a certain extent. Despite multiple dips during the training process, DBAFL eventually converges at an appropriate accuracy level (at around the $50$th second mark). The outcome is consistent with the analysis in Section~\ref{sec:reliability}, as the committee leader identifies local models with low accuracy and assigns a relatively small scaling factor to them. Moreover, when adopting a stricter defense strategy, for example, discarding local models with accuracy lower than the threshold, the resistance of DBAFL towards poisoning attacks is further improved. As shown in Fig.~\ref{fig:compare_attack_poisoning} (c) and (d), when the defense threshold reaches $80\%$ and $90\%$, both the degree and the number of dips are reduced, as the impacts of poisoned local models are mitigated more thoroughly. In addition, with an increased degree of defense, the average test accuracy of the model is also increased slightly (from $46\%$ to $49\%$). This reveals that discarding poisoned local models has no side effects for DBAFL, as the global model learns nothing from the poisoned local models.

When suffering DDoS attacks, the aggregation server in the classic AFL scheme becomes unresponsive to aggregation requests, leading to a lower convergence speed of the global model. As shown in Fig.~\ref{fig:compare_attack_ddos} (a), the convergence time increases from $40$ seconds to $200$ seconds as the DDoS attack traffic increases from $0\%$ to $90\%$. However, the convergence speed of DBAFL is barely affected even when the DDoS attacks traffic is increased to $90\%$. This is due to the periodic election of a random committee leader, which reduces the likelihood of the committee leader being the subject of DDoS attacks.

\begin{framed}
\noindent Result 3: DBAFL is natively resistant to both poisoning and DDoS attacks with the potential to improve reliability further.
\end{framed}

\section{Summary and Future Work}
\label{sec:summary_and_future_works}

This paper offers a blockchain-based asynchronous federated learning scheme with a dynamic scaling factor, aiming to address the challenges faced by FL on IoV networks in terms of learning performance, efficiency, and reliability.
The novel committee-based consensus algorithm in blockchain ensures the reliability of DBAFL with the least cost in communication latency.
In conjunction with the efficient asynchronous aggregation strategy, the dynamic scaling factor assigns reasonable weights to stale local models and improves learning performance for DBAFL.
Extensive experiments conducted on heterogeneous devices validate the advantages of DBAFL in learning performance, efficiency, and reliability.

Future work includes recovering models when nodes go offline unexpectedly, applying DBAFL on non-independent and -identically distributed (Non-IID) datasets, and designing effective strategies to resist other attacks.

\ifCLASSOPTIONcaptionsoff
  \newpage
\fi

\section*{Acknowledgment}

This work was supported in part by the Australian Research Council under grant DP220100983. Many thanks to Wanping Bai for her time and efforts in helping proofread this paper.

\bibliography{refer}{}
\bibliographystyle{unsrt}

\begin{IEEEbiography}[{\includegraphics[width=1in,height=1.25in,clip,keepaspectratio]{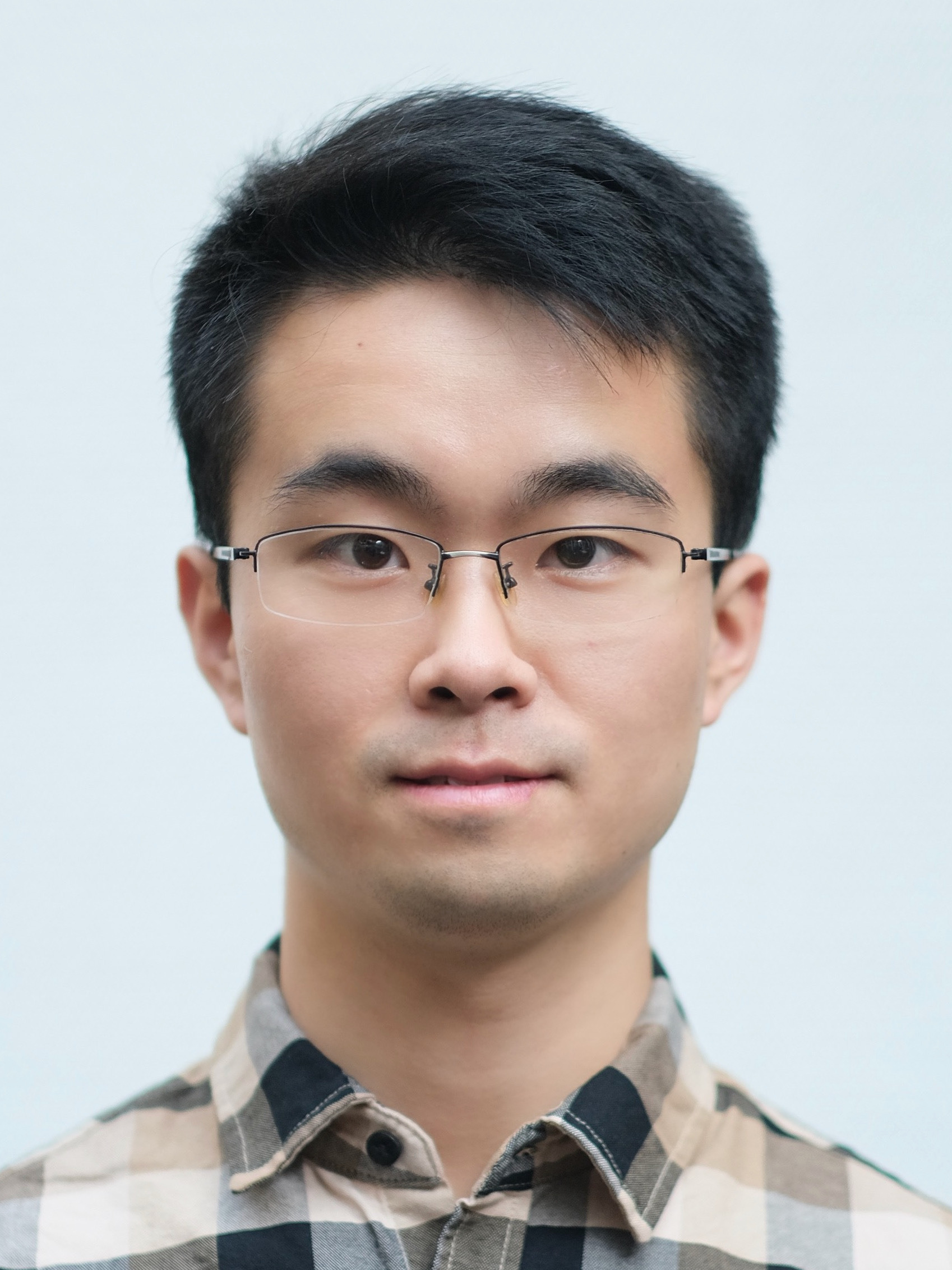}}]{Chenhao Xu}
received a BS degree in Software Engineering in 2018 from Beijing Institute of Technology, China. He is currently pursuing a Ph.D. degree at the School of Information Technology, Deakin University. His research interests include federated learning, blockchain, and IoT.
\end{IEEEbiography}

\begin{IEEEbiography}[{\includegraphics[width=1in,height=1.25in,clip,keepaspectratio]{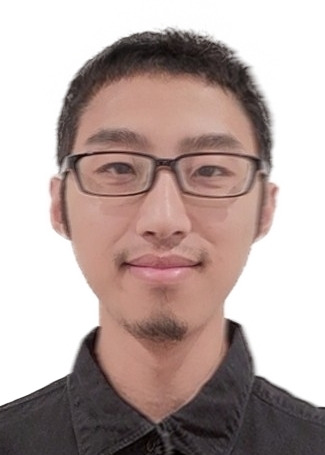}}]{Youyang Qu}
received his B.S. degree of Mechanical Automation in 2002 and M.S. degree of Software Engineering in 2015 from Beijing Institute of Technology, respectively. He received his Ph.D. degree at School of Information Technology, Deakin University in 2019. His research interests focus on dealing with security and customizable privacy issues in Blockchian, Social Networks, Machine Learning, and IoT. He is active in communication society and has served as a TPC Member for IEEE flagship conferences including IEEE ICC and IEEE Globecom.
\end{IEEEbiography}

\begin{IEEEbiography}[{\includegraphics[width=1in,height=1.25in,clip,keepaspectratio]{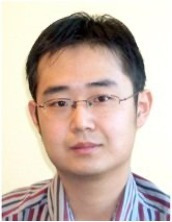}}]{Tom H. Luan}
received the B.Eng. degree from Xi’an Jiao Tong University, China, in 2004, the M.Phil. degree from The Hong Kong University of Science and Technology in 2007, and the Ph.D. degree from the University of Waterloo, Waterloo, ON, Canada, in 2012. He is currently a Professor with the School of Cyber Engineering, Xidian University, Xi’an, China. He has authored/co-authored more than 40 journal papers and 30 technical papers in conference proceedings, and has received one U.S. patent. His research mainly focuses on content distribution and media streaming in vehicular ad hoc networks and peerto-peer networking, and the protocol design and performance evaluation of wireless cloud computing and edge computing.
\end{IEEEbiography}

\begin{IEEEbiography}[{\includegraphics[width=1in,height=1.25in,clip,keepaspectratio]{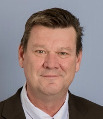}}]{Peter W. Eklund} 
is Professor of AI and Machine Learning at Deakin University’s School of Information Technology. Peter received his PhD from Link\"oping University in Sweden, has an M.Phil from Brighton University in the UK and was a graduate in Mathematics from the University of Wollongong. Peter has over 150 publications and is an elected fellow of the Australian Computer Society.
\end{IEEEbiography}

\begin{IEEEbiography}[{\includegraphics[width=1in,height=1.25in,clip,keepaspectratio]{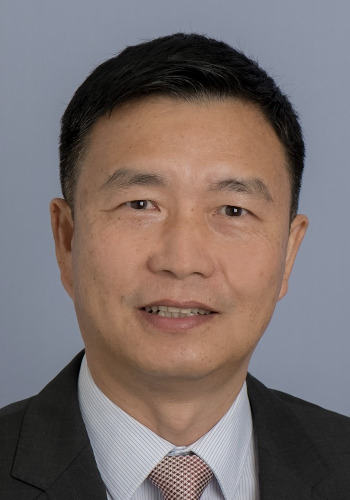}}]{Yong Xiang} (SM'12) received the Ph.D. degree in Electrical and Electronic Engineering from The University of Melbourne, Australia. He is a Professor at the School of Information Technology, Deakin University, Australia. His research interests include distributed computing, cybersecurity and privacy, machine learning and AI, and communications technologies. He has published 7 monographs, over 220 refereed journal articles, and over 100 conference papers in these areas. Professor Xiang is the Senior Area Editor of IEEE Signal Processing Letters, the Associate Editor of IEEE Communications Surveys and Tutorials, and the Associate Editor of Computer Standards and Interfaces. He was the Associate Editor of IEEE Signal Processing Letters and IEEE Access, and the Guest Editor of IEEE Transactions on Industrial Informatics, IEEE Multimedia, etc. He has served as Honorary Chair, General Chair, Program Chair, TPC Chair, Symposium Chair and Track Chair for many conferences, and was invited to give keynotes at numerous international conferences.
\end{IEEEbiography}

\begin{IEEEbiography}[{\includegraphics[width=1in,height=1.25in,clip,keepaspectratio]{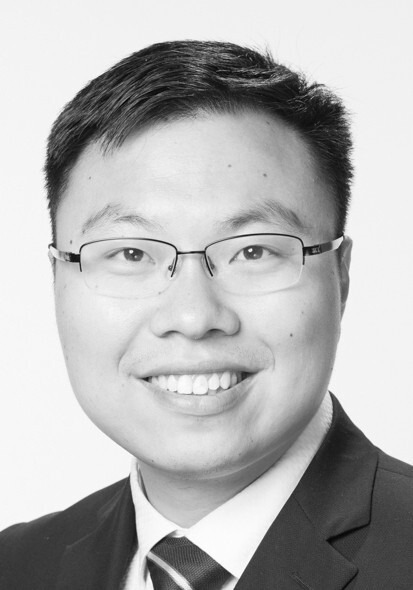}}]{Longxiang Gao} (SM17) received his PhD in Computer Science from Deakin University, Australia. He is currently a Professor at Qilu University of Technology (Shandong Academy of Sciences) and Shandong Computer Science Center (National Supercomputer Center in Jinan). He was a Senior Lecturer at School of Information Technology, Deakin University and a post-doctoral research fellow at IBM Research \& Development, Australia. His research interests include Fog/Edge computing, Blockchain, data analysis and privacy protection.

Dr. Gao has over 90 publications, including patent, monograph, book chapter, journal and conference papers. Some of his publications have been published in the top venue, such as IEEE TMC, IEEE TPDS, IEEE IoTJ, IEEE TDSC, IEEE TVT, IEEE TCSS, IEEE TII and IEEE TNSE. He has being Chief Investigator (CI) for more than 20 research projects (the total awarded amount is over \$5 million), from pure research project to contracted industry research. He is a Senior Member of IEEE.
\end{IEEEbiography}

\end{document}